\documentclass[11pt,a4paper]{article}
\usepackage[hyperref]{acl2021}
\usepackage{times}

\usepackage{microtype}
\usepackage{cleveref}
\usepackage{makecell}
\makeatletter\let\AtPageUpperLeft\relax\makeatother


\makeatletter\let\AtTextUpperLeft\relax\makeatother


\makeatletter\let\ESO@HookII\relax\makeatother
\makeatletter\let\ESO@HookIII\relax\makeatother
\makeatletter\let\AddToShipoutPicture\relax\makeatother

\makeatletter\let\ESO@HookI\relax\makeatother
\makeatletter\let\ESO@griddelta\relax\makeatother
\makeatletter\let\ESO@griddeltaY\relax\makeatother
\makeatletter\let\ESO@gridDelta\relax\makeatother
\makeatletter\let\ESO@gridDeltaY\relax\makeatother
\makeatletter\let\ESO@yoffsetI\relax\makeatother
\makeatletter\let\ESO@yoffsetII\relax\makeatother
\usepackage{pdfpages}
\usepackage{subcaption}
\usepackage{soul}
\usepackage{enumitem}
\usepackage{comment}
\hypersetup{
    colorlinks = false,
    linkbordercolor = {white},
}

\aclfinalcopy 

\newcommand{\name}{$\textit{Target-side Incoherence}$}{}{}
\newcommand{\sname}{$\textit{TSIC}$}{}{}
\newcommand{\tname}{$\textit{InterpretLR}$}{}{}
\newcommand{\mystar}{{\textsuperscript{\fontfamily{lmr}\selectfont$\star$}}}{}{}

\makeatletter
\def\blfootnote{\xdef\@thefnmark{}\@footnotetext}
\makeatother

\crefformat{section}{\S#2#1#3} 
\crefformat{subsection}{\S#2#1#3}
\crefformat{subsubsection}{\S#2#1#3}

\title{How Low is Too Low?\\ A Computational Perspective on Extremely Low-Resource Languages}

\author{Rachit Bansal$^1$ \quad Himanshu Choudhary$^1$ \quad Ravneet Punia$^1$ \\ \textbf{Niko Schenk$^2$$^{\dagger}$ \quad Jacob L Dahl$^3$ \quad Émilie Pagé-Perron$^3$} \\
$^1$~Delhi Technological University \quad $^2$~Amazon Berlin, Germany \quad $^3$~University of Oxford \\
\normalsize
\texttt{\{\href{mailto:rachitbansal2500@gmail.com}{rachitbansal2500}, \href{mailto:himanshu.dce12@gmail.com}{himanshu.dce12}, \href{mailto:ravneet.dtu@gmail.com}{ravneet.dtu}\}@gmail.com} \\
\normalsize
\texttt{\href{mailto:nschenk@em.uni-frankfurt.de}{nikosch@amazon.com}} \\
\normalsize
\texttt{\{\href{mailto:jacob.dahl@gwolfson.ox.ac.uk}{jacob.dahl}, \href{mailto:emilie.page-perron@wolfson.ox.ac.uk}{emilie.page-perron}\}@wolfson.ox.ac.uk}
}

\date{}

\begin{document}
\maketitle
\begin{abstract}
Despite the recent advancements of attention-based deep learning architectures across a majority of Natural Language Processing tasks, their application remains limited in a low-resource setting because of a lack of pre-trained models for such languages. In this study, we make the first attempt to investigate the challenges of adapting these techniques for an extremely low-resource language -- Sumerian cuneiform -- one of the world's oldest written languages attested from at least the beginning of the 3rd millennium BC. Specifically, we introduce the first cross-lingual information extraction pipeline for Sumerian, which includes part-of-speech tagging, named entity recognition, and machine translation. We further curate \tname, an interpretability toolkit for low-resource NLP, and use it alongside human attributions to make sense of the models. We emphasize on human evaluations to gauge all our techniques. Notably, most components of our pipeline can be generalised to any other language to obtain an interpretable execution of the techniques, especially in a low-resource setting. We publicly release all software, model checkpoints, and a novel dataset with domain-specific pre-processing to promote further research.
\blfootnote{Datasets and training subroutines are available at \href{https://linktr.ee/rachitbansal}{\texttt{linktr.ee/rachitbansal}}}
\blfootnote{$^{\dagger}$Work was done prior to joining Amazon at Goethe University Frankfurt}
\end{abstract}

\section{Introduction}

\begin{figure}[h]
\centering
\includegraphics[width=0.49\textwidth]{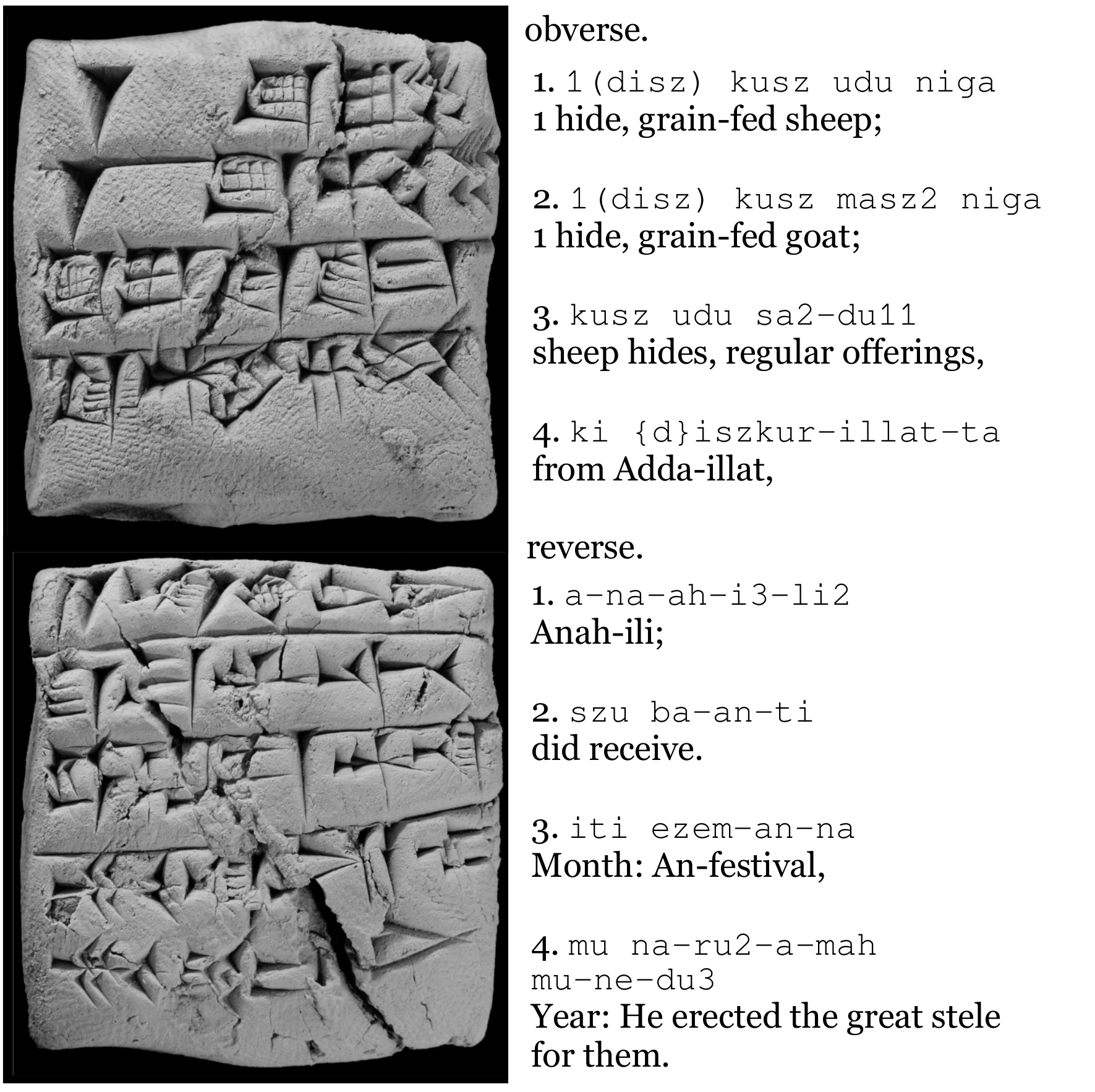}
\setlength{\belowcaptionskip}{-6.25mm}
\vspace{-0.65cm}
\caption{Tablets inscribed with Sumerian cuneiform script, their corresponding digitized transliterations, and human-translated English text for each line.}
\label{tablet}
\end{figure}

Sumerian is one of the oldest written languages, attested in the cuneiform texts from around 2900 BC and possibly the language of even older proto-cuneiform texts from the second half of 4th millennium BC \cite{englund_sum}. Specialists in Assyriology have recently worked to digitize Sumerian scripts, annotate, and translate a part of them to modern-day languages like English and German.

In this work, we attempt to create the first information extraction and translation pipeline for Sumerian. Specifically, we focus on machine translation from Sumerian to English, and sequence labeling tasks of Named Entity Recognition (NER) and Part of Speech Tagging (POS).

Figure \ref{tablet} shows a sample of our raw data where the Sumerian text has been derived from the tablet-inscribed cuneiform script along with its human-interpreted English translations. Creating an annotated corpus for such a language is a tedious task. Thankfully, we obtain our data from openly available sources and corpora, painstakingly annotated and translated by human experts. Yet, for languages like Sumerian, which are not fully-understood by humans themselves, transferring knowledge and patterns to learning algorithms from this limited data becomes extremely difficult. The consequent challenge posed for NER and POS is evident. Lack of annotated data and fuzzy character-level text makes it hard for a model to generalise, irrespective of its size.\\ In case of machine translation, the labeled data is composed of incomplete and short phrase-like sentences, specially on the target-side. This makes the context largely ambiguous. Moreover, we find that for a majority of medieval languages the target-side translated text is highly incoherent with modern-day English language text, making it impossible to use the latter in semi-supervised and unsupervised settings. 

Throughout this study, we elaborate on such challenges induced when working with low-resource languages, and talk about what makes some of these languages like Sumerian `extremely' low-resource. Through extensive experimentation, evaluation, and analysis we further introduce specific algorithms and modifications to work around them.

In all, our contribution is three-fold:
\begin{enumerate}[noitemsep, nolistsep]
\item Building and analyzing a variety of algorithms on the unexplored human-annotated Sumerian dataset for sequence labeling tasks of POS Tagging and NER. (\cref{sec:pos_ner})
\item Introducing the problem of \textit{\name} for low-resource settings and its effect on semi-supervised and unsupervised machine translation (\cref{sec:ssumt}). Further investigating specific modifications and methodologies to cope-up with these constraints. (\cref{sec:mt})
\item Introducing \tname, a generalisable toolkit to interpret low-resource NLP. We apply to additionally study, compare, and evaluate all of the proposed techniques for machine translation and sequence labeling. (\cref{sec:interpret})
\end{enumerate}
Throughout this work, we have conducted human studies and evaluation for our models, in addition to automated metrics. For studying our models with \tname, we have made use of human annotations.

\section{Background}
\subsection{Data}
Sumerian is an ancient language from Iraq that was written using the cuneiform script. While Basque and Turkish display some similarities (split-ergativity, agglutinativity), it is a language isolate \cite{englund_sum}. We have found artifacts dating to around 2900 BC with Sumerian texts inscribed until the first century AD. Most of the Sumerian texts found to this day are administrative in nature as, during the third dynasty of the Ur III Period, the state administration swell to an unprecedented level of activity which was not seen again later in the history of Mesopotamian culture. All through this study, our evaluation sets are composed of Ur III Admin text only and it acts as our in-domain data.

\begin{figure}[h]
\centering
\includegraphics[width=0.48\textwidth, trim={1cm 0.25cm 0 0}, clip]{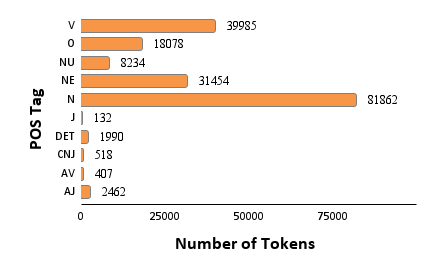}
\setlength{\belowcaptionskip}{-5mm}
\vspace{-0.5cm}
\caption{Composition of the POS tagging dataset. Here, ``NE" stands for named entities, ``O" stands for unstructured words. Other tags are in accordance with ORACC.}
\label{dataposner1}
\end{figure}

\begin{figure}[h]
\centering
\includegraphics[scale=0.5, trim={1cm 0.25cm 0 0}, clip]{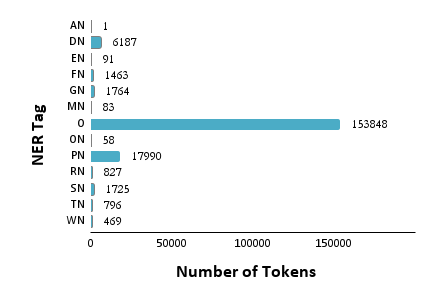}
\setlength{\belowcaptionskip}{-10mm}
\vspace{-0.3cm}
\caption{Composition of our NER dataset. Tags are in accordance with ORACC.}
\label{dataposner2}
\vspace{-0.2em}
\end{figure}

Part of the datasets we used were assembled from the Cuneiform Digital Library Initiative (CDLI)\footnote{\url{https://cdli.ucla.edu}}, Machine Translation and Automated Analysis of Cuneiform languages (MTAAC) project \cite{page-perron-etal-2017-machine}\footnote{\url{https://cdli-gh.github.io/mtaac/}} and The Electronic Text Corpus of Sumerian Literature (ETCSL) dataset \footnote{\url{http://http://etcsl.orinst.ox.ac.uk/}}. CDLI and MTAAC datasets contain the Ur III Administrative (Admin) texts\footnote{The Third Dynasty of Ur is a cultural and temporal period ranging in ${\sim}2112-2004$ BC, in Mesopotamia} which are preserved by the CDLI\footnote{\url{https://github.com/cdli-gh/data}, \url{https://github.com/cdli-gh/mtaac_gold_corpus/tree/workflow/morph/to_dict}}. The MTAAC and ETCSL corpora were both manually annotated for morphology by cuneiform linguistics.\\We divided the data between training and testing sets, and then to reduce the data sparsity, we performed text augmentation using a set of labeled named entities for these sets separately. This increased our combined number of phrases from $25{,}000$ to $48{,}000$, representing our final dataset for sequence labeling. Figures \ref{dataposner1} and \ref{dataposner2} provide the distribution of word tokens in our final pre-annotated dataset. The corpus consists of phrases with lengths ranging from $1$ to $19$ words. These phrases are small since they are translated line by line from the scripts. Around $2{,}500$ phrases were used for testing, while the $45{,}500$ were employed for training purposes.\\
For machine translation, the final dataset summarizes as (i) $10{,}520$ parallel phrases from the Ur III administrative corpus; (ii) $88{,}460$ parallel phrases, all genres combined; and (iii) all monolingual Sumerian data ($1.43$ million phrases). In all cases, phrases are short, generally ranging from 1 to 5-word tokens.

\subsection{Related Work}
Past work aimed at machine translation of Sumerian-English \cite{page-perron-etal-2017-machine, punia_ravneet} have used the minimal bitext upon a variety of general statistical and neural supervised techniques. However, they do not handle the text-level peculiarities any differently than one would do for a high-resource language, thus, often failing to capture context, resulting in poor and inconsistent translations. Techniques, learning algorithms, and architectures that optimally use the vast monolingual data and parallel sentences while keeping in mind the several linguistic limitations are motivated in such a scenario. Thus, we experiment on semi-supervised and unsupervised techniques across the three categories of Data Augmentation \cite{sennrich, he}, Knowledge Transfer \cite{zoph}, and Pre-training \cite{lample, mass}.

In the past, \citet{page-perron-etal-2017-machine} applied statistical models for morphological analysis and information extraction for Sumerian. Although, due to the unavailability of annotated data, these models could not generalise well. \citet{Liu} and \citet{Luo2015UnsupervisedSP} used an unsupervised approach for NER with the help of domain experts and used Contextual and Spelling rules to build the model. They also post-processed their outputs automatically, which enhanced their results. In this work, we thoroughly investigate a wide range of algorithms for these sequence labeling tasks and consequently take a first step towards effective information extraction for Sumerian.



\section{Part of Speech Tagging and Named Entity Recognition} \label{sec:pos_ner}

In this section, we talk about the various algorithms that we investigated to carry out the sequence labeling tasks of POS and NER for Sumerian. The subsequent experimental results are described and discussed in Section \ref{sec:results}.\\
\\
\textbf{Conditional Random Fields}{  }  CRF \cite{inproceedingscrf} is a discriminative probabilistic classifier, which optimises the weights or parameters in order to maximize the conditional probability distribution $P(y\ |\ x)$. They take set of input features (language or domain specific) into account, using the learned weights associated with these features and previous labels to predict the current label. Since CRFs use feature sets (rules) which are language-specific, it makes the model more robust specially for very low-resource languages. In our case we developed domain specific rules with the help of previous studies \cite{Liu, Luo2015UnsupervisedSP} and language experts. A set of these rules are mentioned in the Appendix.\\
\\
\textbf{Bi-directional LSTM}{  }We also experiment across Recurrent Neural Networks (RNNs) to deal with the sequential text input. We employ Bi-LSTM \cite{lstm, lstm2} in particular. As in \citet{huang2015bidirectional}, an additional CRF layer is used for efficient usage of sentence level tag information and past input features by LSTM cells.\\ \\
\textbf{FLAIR}{  }\citet{akbik2018coling} introduced a Contextual String Embedding for Sequence Labeling, FLAIR, which has shown great success for various languages for NER \cite{akbik2019naacl}. We make use of the two distinct properties of its embeddings: (i) training without any explicit notion of words and fundamentally modeling the words as a sequence of characters, and (ii) deriving and using the context from surrounding tokens.\\
We train the bi-directional character language model using the Sumerian monolingual phrases and retrieve the contextual embedding for each word which we then pass into the vanilla Bi-LSTM CRF model.\\ 
\textbf{RoBERTa}{  }We also investigate the transformer-based language model, RoBERTa \cite{liu2019roberta}. The encoder is first pre-trained on our Sumerian monolingual data, and then fine-tuned on our downstream sequence labeling tasks using the labeled data.

\section{Machine Translation} \label{sec:mt}
In this section, we present our experiments for machine translation, primarily focusing on specific data and algorithmic modeling techniques which may be generalised for any extremely low-resource language that may or may not suffer from \name, a phenomenon which we also introduce herein. All results are summarised in Table \ref{nmt}. 

\subsection{Supervised NMT} \label{sec:smt}
In order to create a benchmark for the semi-supervised and unsupervised approaches, we perform supervised machine translation using the limited bitext available (${\sim}10{,}000$ phrases). We perform experiments on a variety of data configurations which are given by:

\begin{enumerate}[noitemsep, nolistsep]
  \item \texttt{UrIIISeg}: Follows the format as present in the original texts provided by Assyriologists and used in the past attempts for Sumerian-English machine translation \cite{page-perron-etal-2017-machine}. It contains only the in-domain Ur III Admin Data with line-by-line translated segments of 1-5 words each, amounting to $10528$ segments.
  \item \texttt{UrIIIComp}: Also contains the in-domain data only, but multiple segments are concatenated together to form complete sentences. The `completeness' of a sentence is ensured using punctuation marks. It comprises of only $4792$ sentences.
  \item \texttt{AllSeg}: Contains out-of-domain Sumerian text segments in addition to in-domain Ur III Admin alone. The additional text varies across a wide range of genres such as literary, lexical, ritual and legal, resulting into a corpus size of $88466$ segments. 
  \item \texttt{AllComp}: Combines the additional features of 2. and 3., thus comprising of a total of $32694$ complete text sentences from all genres. 
\end{enumerate}

We make use of the vanilla transformer encoder and decoder architecture \cite{vaswani} for all our supervised machine translation experiments over these three different bitext configurations. Noting the results as in \ref{nmt}, the \texttt{AllComp} text configuration is used for all other experiments. The computational configurations are mentioned in Section \ref{sec:expset}.

\subsection{Semi-Supervised and Unsupervised NMT} \label{sec:ssumt}
We observed that one of the primary reasons for the lack of success of semi-supervised and unsupervised algorithms for low-resource settings, specially for medieval languages, is \textit{the lack of coherence between monolingual texts in the modern-day corpora to the target-side text in the available parallel corpora}. We refer to this as the \textbf{\name} (\sname) problem for such languages. \\As can be seen from Figure \ref{tablet}, the transliterated English text in our parallel corpora is vastly different from general modern-day English texts. In Sumerian, this is because the text has been human-translated to English on the level of words and small segments due to insufficient knowledge of the language. This results into a contextually distorted English language text, as compared what we see in general corpora. This leads to multiple pitfalls. Most significantly, the colossal monolingual data available for a data-rich target-side language (i.e., English in this case) can no longer be used. This \name\ holds true for most medieval language texts like Sumerian, which makes them `extremely' low-resource.

In this section, we elaborate on the problems caused due to \sname\ and further present our hypothesis on adapting various semi-supervised and unsupervised NMT techniques to deal with them.\\
\\
\textbf{Forward Translation}{  } Back-translation (BT) \cite{sennrich} has been widely used and analysed for NMT across a large set of language pairs. BT uses a reverse model, Sumerian $\leftarrow$ English trained on the existing parallel corpora, when the task is to translate from Sumerian $\rightarrow$ English, and applies it on the target-side monolingual corpus. The synthetic samples thus generated are added to the source-side corpus and a new reverse model is trained on the augmented dataset. It has been shown to outperform its forward counterpart, Forward Translation (FT) \cite{zhang-zong-ft, burlot-yvon-2018-ft}, which instead uses a forward (Sumerian $\rightarrow$ English) model to augment the target-side of the bitext.\\
However, due to \sname, the target-side monolingual data falls into a completely different distribution than what a Sumerian $\leftarrow$ English model is trained on. Using back-translation in such a scenario results into a poor source-side augmentation, doing more harm than good. 
Keeping this in mind, we rely on forward-translation (FT), thus using the Sumerian monolingual text.

We divide the Sumerian monolingual data into $8$ shards, each containing ${\sim}100{,}000$ monolingual \texttt{AllComp} sentences each. The FT process takes place for each shard and the Transformer model is trained after each shard is forward-translated.

Large scale studies \cite{edunov, wu-etal-bt-ft} have shown the heavy dependency of BT and FT on aspects like sampling methods and the amount of parallel data. The performance with non-MAP (where, MAP stands for \textit{maximum a posteriori}) estimation methods like Nuclear Sampling \cite{holtzman-k} and Beam Search with Noise improves almost-linearly with the amount of bitext, and thus, for low-resource settings (${\sim}80{,}000$ sentence pairs), MAP methods have been shown to give better results. This was also observed in our experiments and the reported results are obtained using Beam Search (\cref{sec:expset}).\\
\\
\textbf{Cross-Lingual Language Model Pre-training}{  } We further make use of XLM \cite{lample} to carry out a wide range of experiments for both unsupervised and semi-supervised fine-tuning techniques. Considering the lack of original target monolingual text due to \sname, the following target data configurations were used for pre-training the XLM:

\begin{enumerate}[noitemsep, nolistsep]
    \item \texttt{WMT:} Ignoring the lack of coherence between general English texts and the evaluation + training texts, to compose the entire target monolingual data with the WMT '18 English Texts. Amounts to a total of $20$M sentences.
    \item \texttt{Orig:} Composed of all the English side texts in \texttt{UrIIISeg}, \texttt{UrIIIComp}, \texttt{AllSeg} and \texttt{AllComp} combined. Contains only ${\sim}60{,}000$ sentences. 
    \item \texttt{Mixed:} This consists of all of 2. and as many sentences as 1. through which the net size of the corpus equalizes the Sumerian monolingual, i.e., $1.5$M.
\end{enumerate}

In the pre-training phase, we perform various experiments over different combinations of MLM and TLM. It is further fine-tuned on a denoising auto-encoding objective for Unsupervised while cross-reference machine translation objective over the parallel data for semi-supervised training. BT steps are also performed in both cases.\\
\\
\textbf{Data Augmentation}{  } In order to further reduce the effect of \sname\ on the model performance and to allow the model to attend to a larger and more diverse volume of target text during pre-training, we make use of the following data augmentation techniques:

\begin{enumerate}[noitemsep, nolistsep]
    \item \texttt{BERT:} Replacing words by the spatially closest words measured by Cosine Similarity in BERT \cite{devlin2018bert} Embeddings, with a threshold of 0.8.
    \item \texttt{WordNet:} Replacing words with WordNet \cite{wordnet} synonyms.
    \item \texttt{CharSwap:} Introduces certain character-level perturbations in the text by substituting, deleting, inserting, and swapping adjacent character tokens.
\end{enumerate}

\begin{figure}[h]
\centering
\includegraphics[scale=0.45,  trim={1cm 0 0 1cm}, clip]{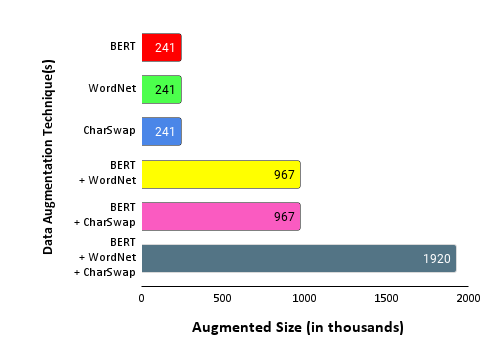}
\setlength{\belowcaptionskip}{-5mm}
\vspace{-0.45cm}
\caption{Effective size of the target monolingual corpora with different combinations of augmentation.}
\label{daug}
\end{figure}

Different combinations of these techniques have been used to augment the \texttt{Orig} type target monolingual data. The resultant target-side corpora sizes are summarised in Figure \ref{daug}.

\section{Experimental Setup} \label{sec:expset}

\begin{table*}[h]
\parbox{.64\linewidth}{
{\renewcommand{\arraystretch}{1.10}
    \centering
    \begin{tabular}{| c | c c c c |  }
     \hline
     \hline
     Technique& S& \makecell{US}& \makecell{SS}& \makecell{HE}\\
     \hline
     \multicolumn{5}{l}{\textit{Vanilla Transformer}} \\
     \hline
     \texttt{UrIIISeg}& 36.32& & &2.202 \\
     \texttt{UrIIIComp}& 33.45& & &2.242 \\
     \texttt{AllSeg}& 37.01& & &2.360 \\
     \texttt{AllComp}& 42.23& & &2.431 \\
     +3$\times$FT\mystar & & & 41.98& 2.358 \\
     \textbf{+5$\times$FT}& & & \textbf{44.14} &\textbf{2.504} \\
     +7$\times$FT& & & 42.95 &2.367 \\
     \hline
     \multicolumn{5}{l}{\textit{XLM}} \\
     \hline
     MLM, \texttt{Orig}& & 4.49& 15.04& \\
     MLM + TLM, \texttt{WMT}& & 0.94& --& \\
     \texttt{Mixed}& & 13.08& 21.23& 1.104, -- \\
     \texttt{Orig}& & 12.73& 24.64& 1.294, -- \\
     \hline
     \multicolumn{5}{l}{\textit{XLM + Data Augmentation}} \\
     \hline
     \texttt{BERT} & & 13.06& 29.50& 1.320, 1.704 \\
     \texttt{WordNet} & & 13.08& 28.57& 1.269, 1.690 \\
     \texttt{CharSwap} & & 12.92& 29.04& \\
     \texttt{BERT+WordNet} & & 13.34& 26.57& 1.460, 1.666 \\
     \texttt{\makecell{BERT+CharSwap\\+WordNet}}& &13.23 & 30.10& -- , 1.757 \\
     \hline
     \end{tabular} \\
    \caption{Sumerian-English Machine Translation. Here, S: Supervised, US: Unsupervised, SS: Semi-Supervised and HE: Human Evaluation. Each of the available values for the first three columns (BLEU) is compared with a value under HE (out of $3$). \mystar Number of shards used for FT.}
    \label{nmt}
    }}
\hfill
\parbox{.33\linewidth}{
{\renewcommand{\arraystretch}{1.15}
    \centering
    \begin{tabular}{|c|c|}
    \hline
    & F1-Score \\ 
    \hline
    HMM & 0.815 \\
    \hline
    \textbf{\makecell{Rules + \\ CRF}}  & \textbf{0.991} \\ 
    \hline
    \makecell{Bi-LSTM + \\ CRF} & 0.763 \\ 
    \hline
    FLAIR & 0.499 \\
    \hline
    RoBERTa & 0.949 \\ 
    \hline
    \end{tabular}
    \caption{POS Tagging for Sumerian. CRF with rules outperform large models like FLAIR and RoBERTa.}
    \label{table:POSResults}
    
    \vspace{0.5cm}
    
    \begin{tabular}{|c|c|}
    \hline
    & F1-Score \\
    \hline
    HMM & 0.656 \\
    \hline
    \makecell{Rules + \\ CRF} & 0.913 \\
    \hline
    \makecell{Bi-LSTM + \\ CRF} & 0.775 \\ 
    \hline
     FLAIR & 0.187 \\
    \hline
    \textbf{RoBERTa} & \textbf{0.953} \\
    \hline
    \end{tabular}
    \caption{NER for Sumerian. RoBERTa performs best among others. Due to high character-level noise, FLAIR fails to generalise well.}
    \label{table:NREResults}
}}
\end{table*}

All our experiments have been implemented in PyTorch, except for the Bi-LSTM and CRF which were done in Tensorflow. In addition to this, we used FairSeq \cite{ott2019fairseq}, FLAIR \cite{akbik-etal-2019-flair}, HuggingFace Transformers \cite{HuggingFace}, and Open-NMT \cite{opennmt} frameworks in Python. Nvidia Apex was used for memory optimisation using fp-16 training. Experiments related to Bi-LSTM, CRF, vanilla transformers, and FT were performed on a single $8 GB$ Nvidia GeForce RTX 2070 GPU, while the pre-training and fine-tuning of FLAIR, RoBERTa, and XLM on various data configurations were performed on $2$ $16$ GB Nvidia V100 GPUs. We used development sets to tune the hyper-parameters for all our models, especially those for POS and NER. For RoBERTa and vanilla transformer, $N=6$ encoder layers with $h=16$ attention heads were used, while $N=4$ and $h=12$ was used for XLM. A beam-size of $5$ was used for our FT experiments. Adam \cite{adam} optimiser with a learning rate of $0.001$, $\beta_1=0.90$, $\beta_2=0.98$ and a decay factor of $0.5$ was used. Additional regularisation was done via Dropout and Attention Dropout (wherever applicable) layers with $p_{drop}=0.1$. We used a batch size of $32$ or $64$ and an early stopping criteria based on the validation loss.

\section{Results and Analysis} \label{sec:results}

\textbf{Sequence Labeling}{  } Tables \ref{table:POSResults} and \ref{table:NREResults} represent the metric scores of our different models for POS and NER tasks, respectively. CRF with domain-specific rules gives the best F1-score for the POS tagging task, even better than the complex RoBERTa and FLAIR language models which are the current state-of-the-art techniques for most languages. The prevalence of distorted words and short phrases in the corpora makes context learning difficult, although the domain-specific rules help learn short-term dependencies by learning feature weights.

RoBERTa performs well for both of the tasks, while being the best among others for NER ($95.37$ F1 score). To make the most out of the limited vocabulary and noisy text, we used Byte-Level BPE \cite{radford2019language} to train the language model and further fine-tuned it on our POS and NER dataset with a batch size of $128$. We also tried FLAIR language model across various word embeddings (character, Word2vec, FastText, GloVe) along with an additional CRF layer for both of the tasks. Although a high precision is observed using this approach, the F1 scores is seen to be significantly low due to low recall. In addition to the F1 metric we also conducted human evaluation by language expert for the best performing models, out of randomly selected $76$ ($496$ words) phrases, only $8$ and $6$ words were misclassified by NER and POS models, giving an error of $1.20$ and $1.61 \%$, respectively.\\
\\
\textbf{Machine Translation}{ }Table \ref{nmt} summarises our results for all supervised, semi-supervised, and unsupervised techniques. Forward translation on vanilla transformer outperforms all other techniques by at least $2$ BLEU. The variation of its performance with more monolingual source text is shown. The superior performance of \texttt{AllComp} over the other configurations in vanilla transformer signifies the value of both context and out-of-domain data together. Even though the XLM-based models show lower performance, it could be attributed to the lesser number of encoder layers and attention heads used for them. What is interesting to note, though, is the variation of its performance across various training strategies. We experiment across MLM and TLM (+ MLM) initialization for XLM, where the latter comfortably outperforms the former. We do not test with random initialization and CLM, following up from the conclusions made for NMT in \citet{lample}. Pre-training the XLM on augmented target-side text works surprisingly well. We note that using pre-training on \texttt{BERT} and \texttt{WordNet} augmentations results in better Unsupervised performance while introducing \texttt{CharSwap} improves the semi-supervised models. The human evaluation presented in the table was made by three Assyriologists, who rated $100$ output examples for each model, on a scale of $3$. A pairwise inter-annotator agreement of $0.673$ (Cohen's Kappa) was observed.\footnote{Elaborate evaluation criteria mentioned in the Appendix.}

\section{Interpretability Analysis} \label{sec:interpret}
\begin{table*}[t]
    \begin{subfigure}{\textwidth}
        \centering
        \includegraphics[scale=0.55]{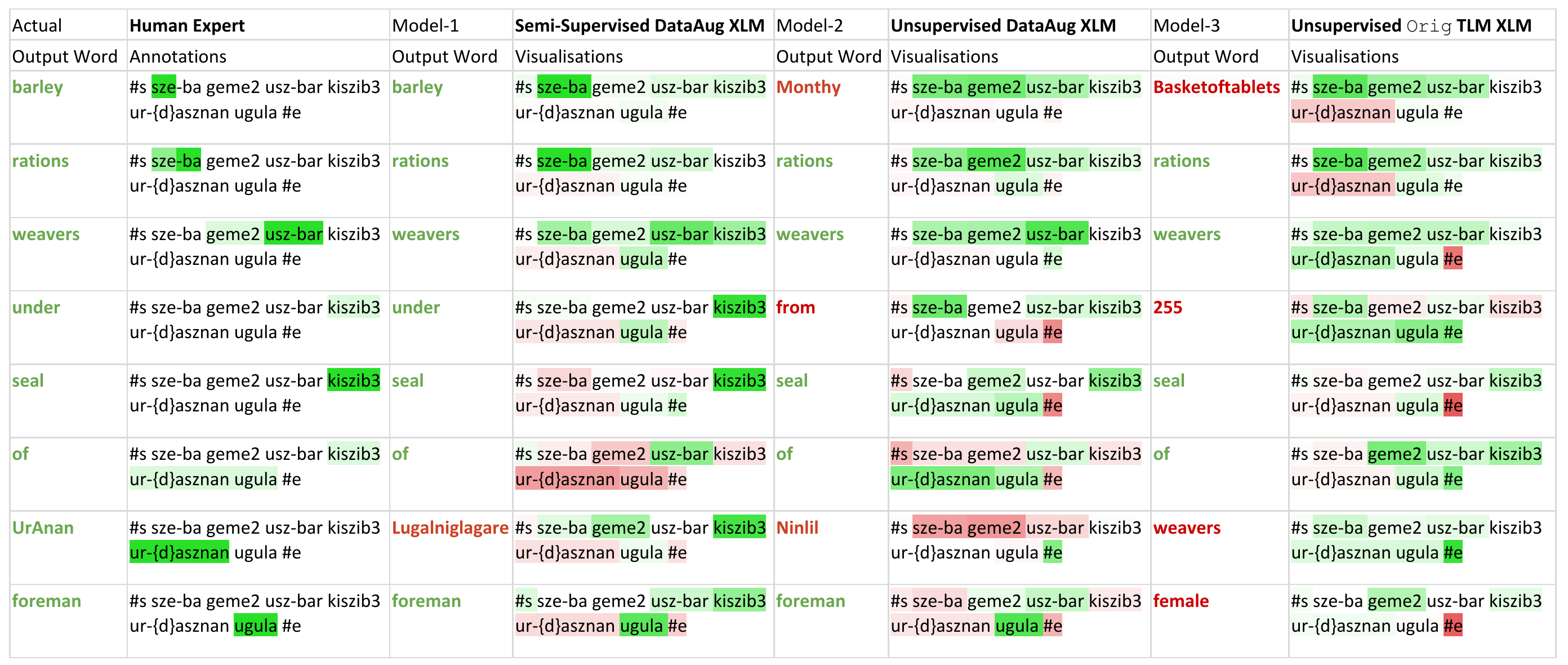}
        \caption{MT- Selected output tokens for Sumerian Input text of ``sze-ba geme2 usz-bar kiszib3 ur-{d}asznan ugula", which translates to ``barley rations of the female weavers under seal of UrAnan the foreman".\protect\footnotemark}
        \label{fig:mt_highlights}
    \end{subfigure}
    \begin{subfigure}{\textwidth}
        \centering
        \begin{subfigure}{0.45\textwidth}
            \centering
            \includegraphics[width=\textwidth]{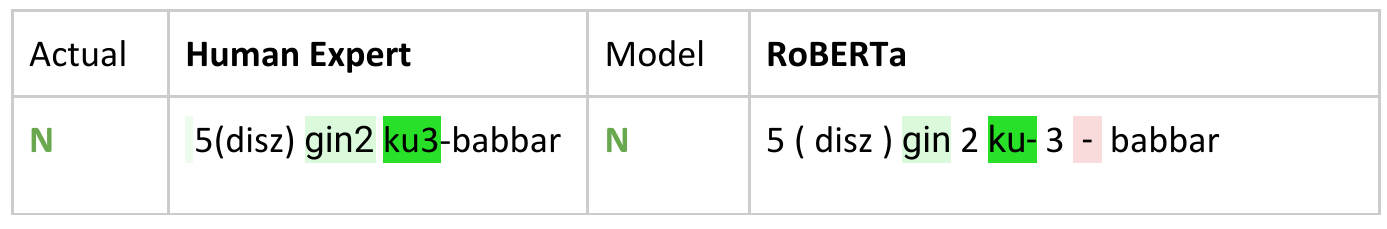}
            \caption{POS- With tagged word ``ku3-babbar"}
            \label{fig:pos_1}
        \end{subfigure}
        \begin{subfigure}{0.45\textwidth}
            \centering
            \includegraphics[width=\textwidth]{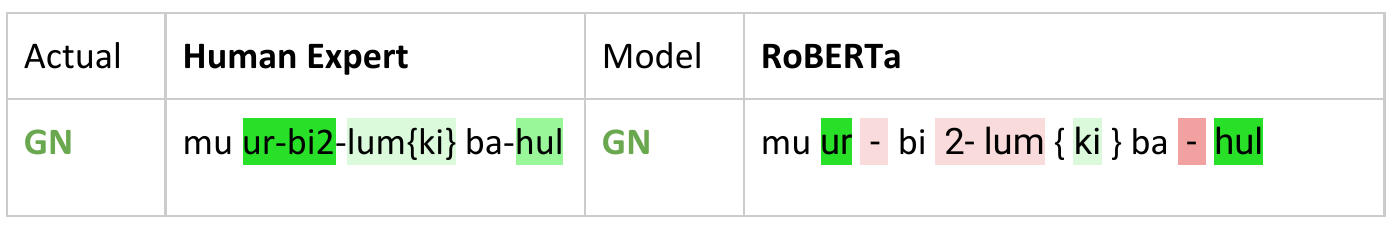}
            \caption{NER- With tagged word ``ur-bi2-lum{ki}"}
            \label{fig:pos_2}
        \end{subfigure}
    \end{subfigure}
    \caption{Highlighted attributions for randomly selected examples. \textcolor{green}{\textbf{Green}} and \textcolor{red}{\textbf{Red}} represent correct and wrong predictions, respectively, while \sethlcolor{green} \hl{Green} and \sethlcolor{red} \hl{Red} highlights represent positive and negative attributions, respectively.}
    \vspace{-1em}
\end{table*}

Oftentimes in case of Deep Learning Architectures, metric scores like Accuracy, F1 and BLEU are unable to portray the true behavior of the models. For languages like Sumerian, the human-understanding itself is scarce. Visualizing the representations and correlations made by the model could provide insights into which elements of the context can give additional information to support semantic analysis of the terms. Thus, we herein introduce a generalisable interpretability toolkit, \tname, to interpret algorithms for \textbf{L}ow-\textbf{R}esource NLP and further apply it for the aforementioned tasks and models. 

\tname\ is primarily aimed at fabricating attribution saliency maps, i.e., tracing back the model output so as to assign an importance score to each input token, based on its `influence' on that output. We do this using two kinds of interpretability techniques-- gradient-based \cite{integratedg, saliency, deep_lift}, and perturbation-based \cite{occlusion, shapley}. 

Due to the inherently discrete nature of natural language text, the starting point for all our approaches is the embedding of the input sentence across the model to interpret. Most of our analysis is done for the encoder of the network architecture, thus analyzing the effect of different pre-training and fine-tuning techniques on how the model eventually represents the language attributes. We use the word `Attribution' as a better-defined substitute for the `Influence' measure of an input span of text on the output. \\A part of our visual analysis is shown and elaborated here, while a complete analysis with all our models and layer-wise heat-maps is presented in the Appendix.

In Table \ref{fig:mt_highlights}, we apply \tname\ on $3$ different configurations of XLM for a randomly chosen sentence from NMT's evaluation set. A human expert was asked to annotate the source sentence in accordance with the expected reference for each output token in the actual English translation, as shown in the first column. The highlighted visualizations for each of the $3$ models were obtained using Integrated Gradients \cite{integratedg} across the three input embeddings- token, position, and language. A lot of interesting observations could be made from these attributions.\\Firstly, the named entity in the sentence \textit{ur-\{d\}asznan} (\textit{UrAnan}) has been wrongly translated by all the three models. Although this behavior is expected (learning the context of a named entity is extremely difficult without excessive supervision around the same, which is largely absent our training text) the models even largely fail to attend to the right words in the input. \\Secondly, words like \textit{rations}, \textit{weavers} and \textit{seal} which appear frequently in the parallel Ur III Admin corpora and have a contextual meaning attached to them, are translated perfectly by the models, this property is observed among these models in general. Even the unsupervised models that do not have access to the one-to-one mapping of the translation during training manage to infer these words from the appropriate context. It can be assumed that they learn the right representations of such tokens. But at the same time, there are instances like \textit{sze-ba} (\textit{barley}), which the two unsupervised models rightly refer to but do not give the right translations, which thus is a direct result of the absence of supervision. \\Lastly, English words like \textit{under}, \textit{of} and \textit{from} do not have any direct translations in Sumerian and are mostly inferred from the context, even by the human annotators. At such places, again, supervision might play a critical role as in the $4^{th}$ row of Table \ref{fig:mt_highlights}. There are also instances like the $6^{th}$ row where the supervised model fails to attend to the right words, and the correct output word could very well be out of memorisation.\footnotetext{The left-out tokens were rightly predicted by all the three models, with almost the same attributions.}

Tables \ref{fig:pos_1} and \ref{fig:pos_2} represent visualizations for two randomly selected phrases for our sequence labeling tasks, indicating the attributions for each sub-word for tagging the corresponding target word with their predicted labels. It can be observed from Table \ref{fig:pos_1} that word \textit{gin} (\textit{unit}) and sub-word \textit{ku}, are contributing to the attribution score positively, depicting positive model attribution to tag \textit{ku3-babbar} as a Noun (N), whereas in Table \ref{fig:pos_2} the sub-words \textit{ur}, \textit{hul} and \textit{ki} are contributing \textit{ur-bi2-lum\{ki\}} to be tagged as the label GN (Geographical Name). As observed from the corresponding human annotation, \textit{ur} and \textit{ki} are the most associated for Geographical names and GNs are mostly followed by a verb part, which is \textit{hul} (\textit{destroy}) in this case. It can thus be inferred that RoBERTa identifies this correspondence well and makes the decision accordingly.


\section{Conclusion}
In this work, we introduced the first information extraction and translation pipeline for Sumerian cuneiform. We first undertook the tasks of POS Tagging and NER, where we observed that \textit{deeper is not necessarily better}. A simple CRF model with well-defined rules outperformed the large language model RoBERTa for POS Tagging. Further, for machine translation we overcame unprecedented challenges pertaining to lack of in-domain text, sparse sentence formation, and incoherence. We found that using out-of-domain text along with specific data-augmentation can have huge impacts in a low-resource setting. All components of this work are generalisable to other low-resource languages, including \tname, and we open way to future research in this direction.



\section*{Acknowledgments}
The authors would like to acknowledge the use of the University of Oxford Advanced Research Computing (ARC) facility in carrying out this work (http://dx.doi.org/10.5281/zenodo.22558). We would like to thank our collaborators at the Cuneiform Digital Library Initiative (CDLI; https://cdli.ucla.edu) and from the Machine Translation and Automated Analysis of Cuneiform Languages (MTAAC; https://cdli-gh.github.io/mtaac/). We would also like to thank CDLI and the Electronic Text Corpus of Sumerian Literature (ETCSL) for providing the data for our experiments. This work was partly undertaken during the Google Summer of Code (GSoC) program, 2020. CDLI has been supported by GSoC, where aspects of machine translation have been addressed by several students since 2018. We are thankful to Ilya Khait and Bertrand Lafont for their assistance with the human evaluations for machine translation and sequence labeling. We are grateful to Orhan Firat for insightful discussions and multiple rounds of reviews during the pre-submission mentoring phase of ACL SRW that greatly shaped this manuscript.

\bibliographystyle{acl_natbib}
\bibliography{anthology,acl2021}
\clearpage
\appendix






\maketitle

\section{Detailed Evaluation and Analysis}

\begin{figure}[h]
\hspace{-1cm}
\includegraphics[width=0.575\textwidth]{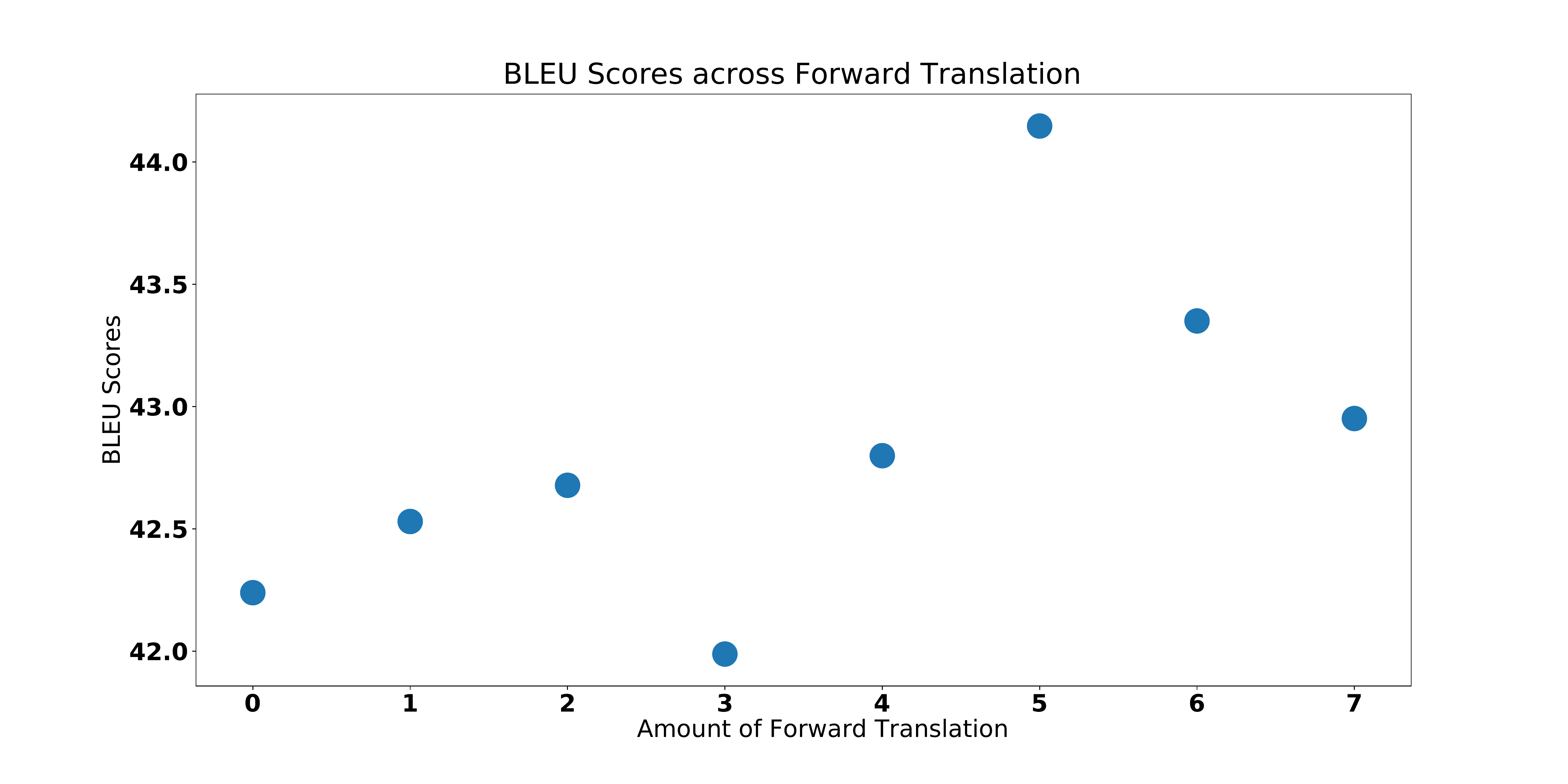}
\caption{}
\label{BT}
\end{figure}

Forward Translation with Vanilla Transformer gave the best results for Sumerian-English Neural Machine Translation. Figure \ref{BT} shows the variation of the BLEU score with the amount of source monolingual data used. Here, the X-Axis represents the number of shards used, with each shard consisting of 80K sentences. It can be observed that the translation accuracy is not linear with the amount of text used.

\begin{figure}[h]
    \centering
    \begin{subfigure}{0.5\textwidth}
        \includegraphics[scale=0.375]{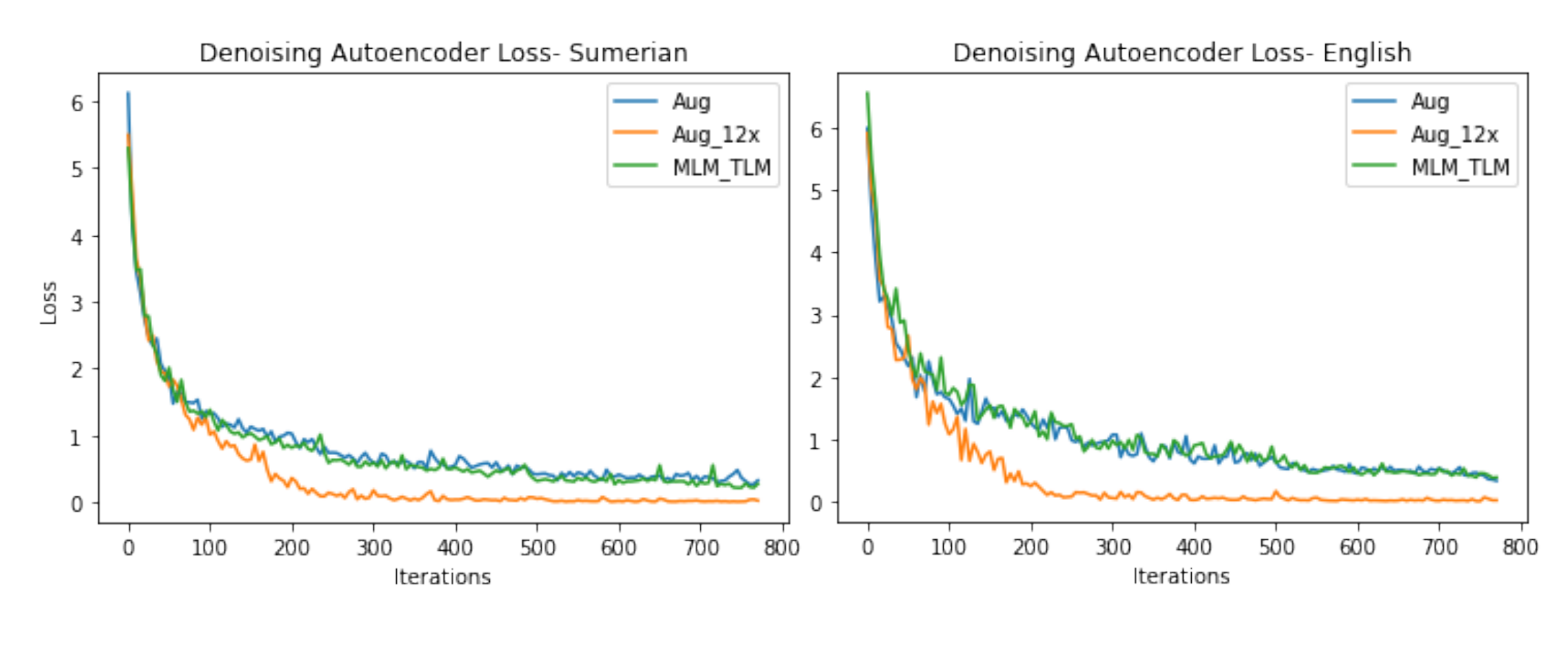}
        \caption{Denoising Auto-encoder Loss (AE Loss) variation across the 1st Epoch}
        \label{AE}
    \end{subfigure}
    \begin{subfigure}{0.5\textwidth}
        \includegraphics[scale=0.375]{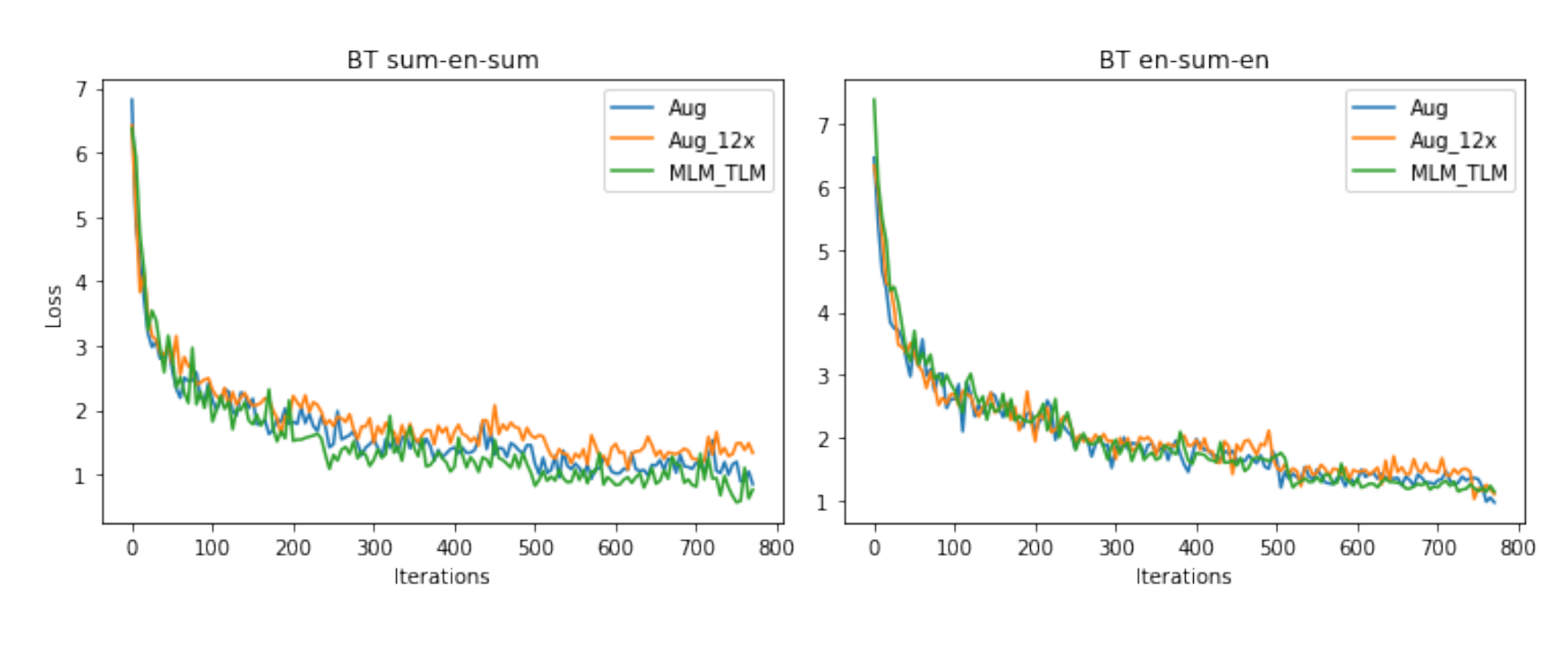}
        \caption{Back Translation Loss variation in XLM across the 1st Epoch}
        \label{BT_2}
    \end{subfigure}
    \begin{subfigure}{0.5\textwidth}
        \includegraphics[scale=0.375]{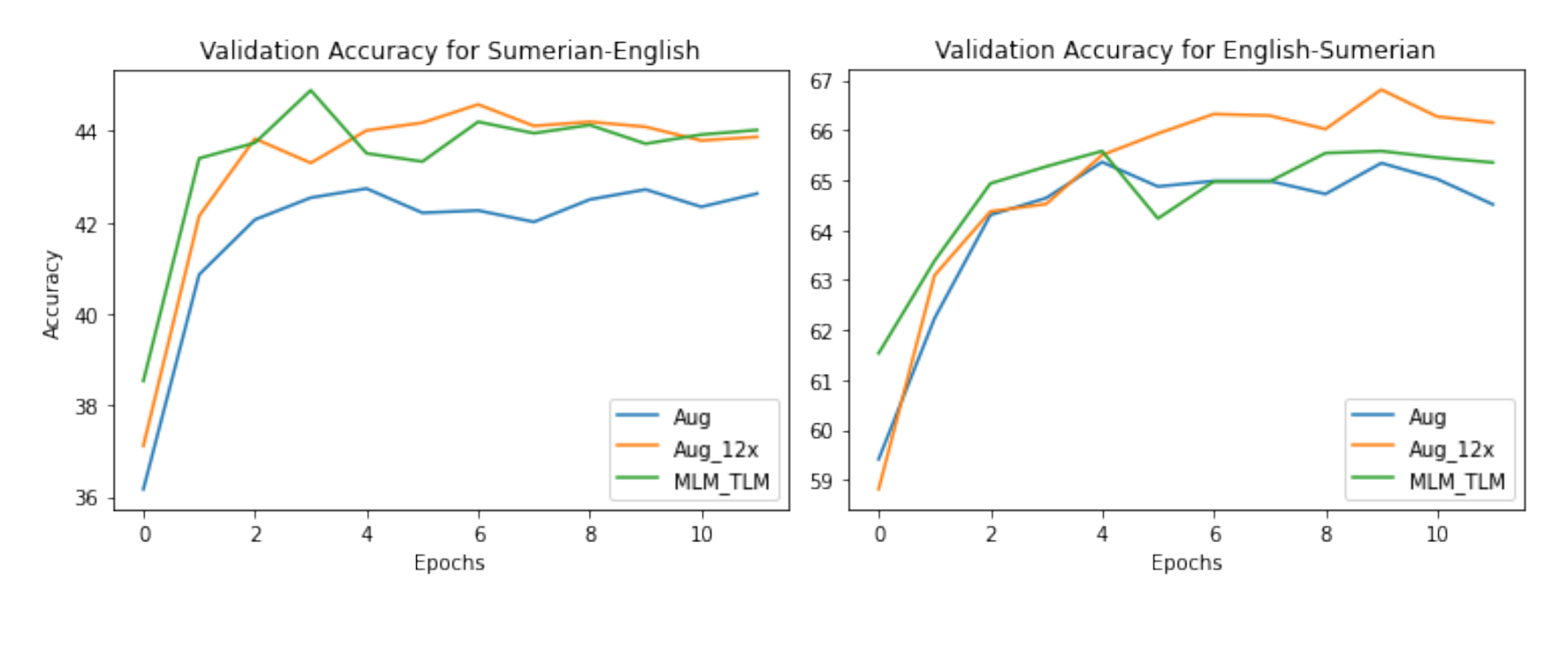}
        \caption{MT accuracy across a number of training epochs}
        \label{val}
    \end{subfigure}
    \caption{Quantitative comparison of various models during Unsupervised MT fine-tuning}
    \label{graphs}
\end{figure}

Figure \ref{graphs} shows the variation of several performance metrics during the Unsupervised fine-tuning of various XLM configurations. The comparison is made between XLM pre-training without any data augmentation (MLM\_TLM), with one augmentation (Aug) and with all three augmentations (Aug\_12x). It can be seen from Figure \ref{AE} that an XLM pre-trained on the Aug\_12x configuration converges the fastest among the others, in terms of the main Denoising Auto-encoding Loss. It can also be observed that the curve corresponding to this configuration is much smoother than the others, which shows a positive regularizing effect of a better weight initialisation (through appropriate pre-training). A similar pattern is observed for the validation accuracy across the epochs as shown in Figure \ref{val}, although, the trend of Back Translation loss remains mostly inseparable for the three configurations. 

Table \ref{Human_results} depicts the net percentage error found by an human expert on the POS and NER results for the entire evaluation set across the best performing model. Table \ref{tabel_4:POSResults} and \ref{tabel_3:NREResults} represents the detailed results of POS and NER models. It can be observed from the tables, that although CRF and RoBERTa models gave the best results, FLAIR language model along with character embeddings also gave high precision for both of the tasks.   

\begin{table}[h]
\centering
\begin{tabular}{|l|l|l|}
    \hline
    & \makecell{POS error\\(in \%)} & \makecell{NER error\\(in \%)} \\ 
    \hline
    Human Evaluation & 1.61 & 1.20 \\
    \hline
    \end{tabular}
    \caption{Human Evaluation for POS and NER}
    \label{Human_results}
\end{table}

\begin{table}[h]
\begin{tabular}{|c|c|c|c|}
    \hline
    & \multicolumn{3}{c|}{Part of Speech Tagging} \\ 
    \cline{2-4} 
    & Precision & Recall & F1-Score \\
    \hline
    HMM & 0.857 & 0.794 & 0.815 \\ 
    \hline
    \makecell{Rules +\\ CRF} & \textbf{0.994} & \textbf{0.989} & \textbf{0.991} \\ \hline
    \makecell{BBi-LSTM \\ + CRF} & 0.852 & 0.710 & 0.7631 \\ 
    \hline
    \makecell{FLAIR} & 0.9323 & 0.4766 & 0.4999 \\ 
    \hline 
    \makecell{RoBERTa} & 0.9500 & 0.9489 & 0.9495 \\ 
    \hline
\end{tabular}
\caption{POS Tagging Models for Ur III Sumerian Text}
\label{tabel_4:POSResults}
\end{table}

\begin{table}[h]
\begin{tabular}{|c|c|c|c|}
    \hline
    & \multicolumn{3}{c|}{Named Entity Recognition} \\ 
    \cline{2-4} 
    & Precision & Recall & F1-Score \\ 
    \hline
    HMM & 0.810 & 0.599 & 0.656 \\
    \hline
    \makecell{Rules +\\ CRF} & 0.916 & 0.910 & 0.913 \\
    \hline
    \makecell{Bi-LSTM \\ + CRF} & 0.864 & 0.704 & 0.775 \\
    \hline
    \makecell{FLAIR} & \textbf{0.9562} & 0.1817 & 0.1873 \\ 
    \hline
    RoBERTa & 0.9540 & \textbf{0.9534} & \textbf{0.9537} \\ 
    \hline
\end{tabular}
\caption{NER Models for Ur III Sumerian Text}
\label{tabel_3:NREResults}
\end{table}

\section{Extended Interpretations}
Here we present the interpretability analysis across a larger set of models and visualisations. We use and compare the different algorithms across layer-level, gradient-based, and perturbation-based techniques to obtain the attributions.

\begin{figure}[h]
    \centering
    \begin{subfigure}{0.5\textwidth}
          \includegraphics[scale=0.35]{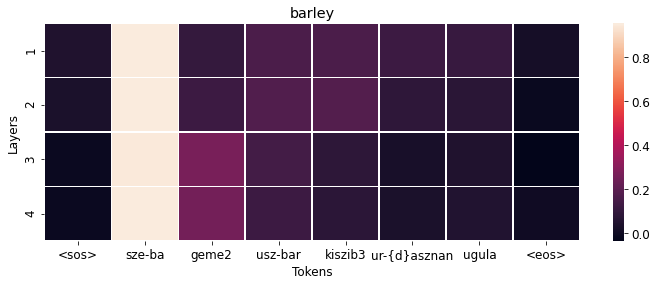}
    \end{subfigure}
    \begin{subfigure}{0.5\textwidth}
          \includegraphics[scale=0.35]{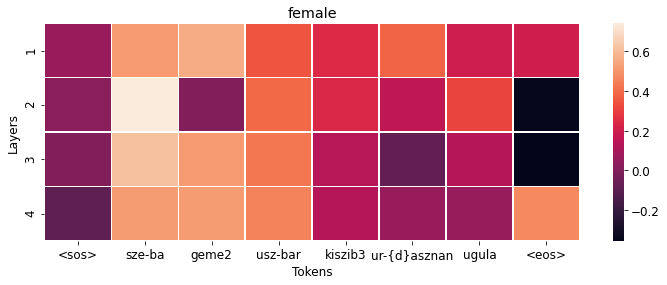}
    \end{subfigure}
    \begin{subfigure}{0.5\textwidth}
          \includegraphics[scale=0.35]{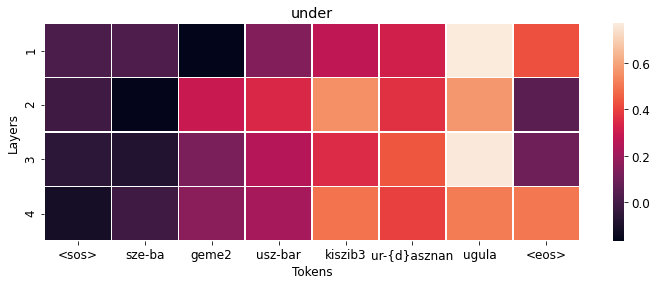}
    \end{subfigure}
    \caption{Layer Conductance across MHSA Layers}
    \label{LC}
\end{figure}

Figure \ref{LC} visualises the Multi-head Self Attention (MHSA) using Layer Conductance Dhamdhere, Sundararajan, and Yan 2018) across the 4 encoder layers we employ in XLMs\footnote{The supervised version of the augmented pre-training is used here.}. The first two output tokens \textit{barley} and \textit{female} are known to be one-on-one mapping between the input words of \textit{sze-ba} and \textit{geme2} respectively. While the third output token \textit{barley} is not a direct translation and is needed to be inferred from context.

\begin{figure*}[htb]
    \centering
    \vspace{-0.8cm}
    \begin{subfigure}{\textwidth}
        \begin{subfigure}{0.33\textwidth}
          \includegraphics[scale=0.25, trim=0cm 0cm 0cm 0cm]{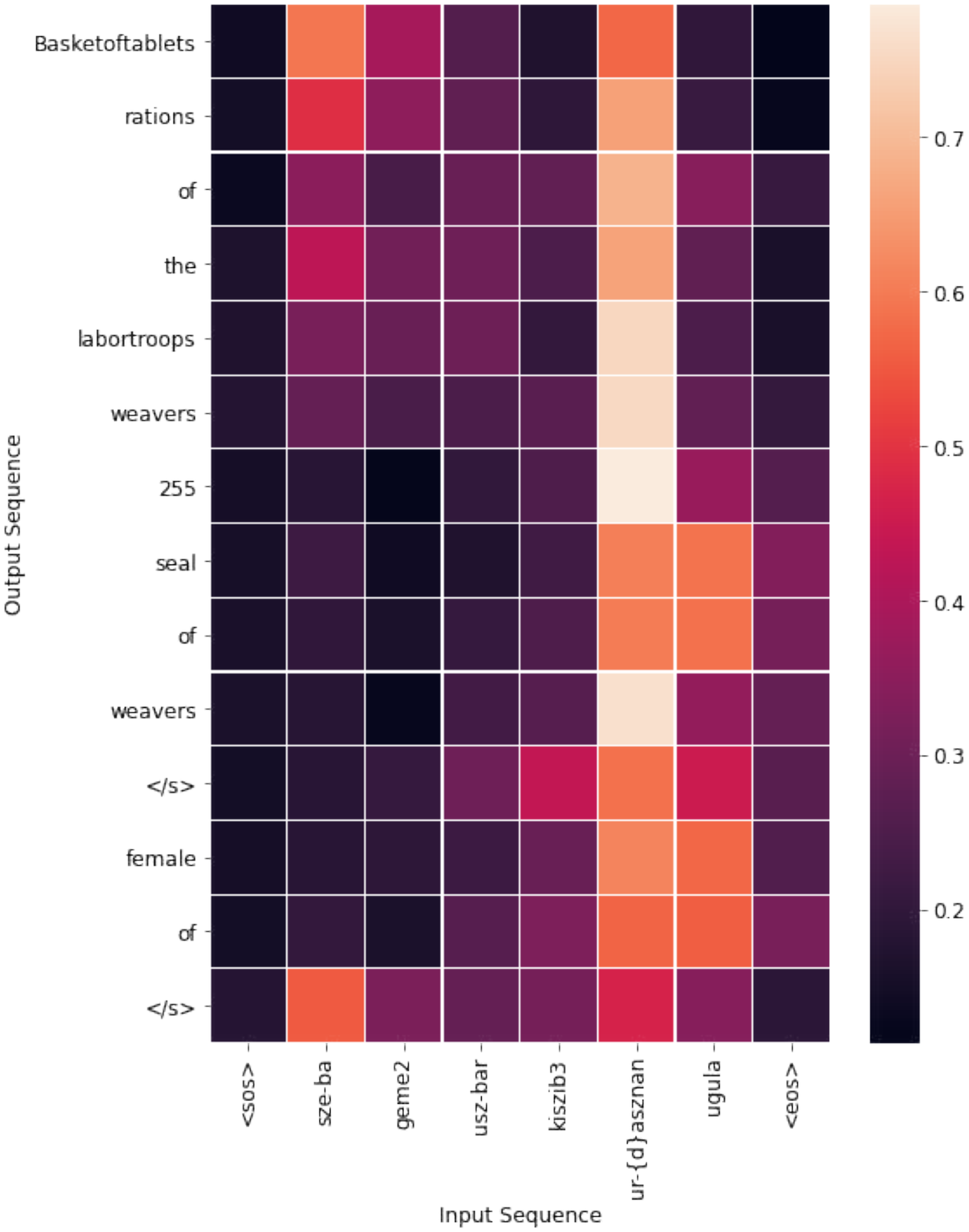}
          \label{fig:tlmsal}
        \end{subfigure}%
        \hspace{0.05cm}
        \begin{subfigure}{0.33\textwidth}
          \includegraphics[scale=0.25, trim=0cm 0cm 0cm 0cm]{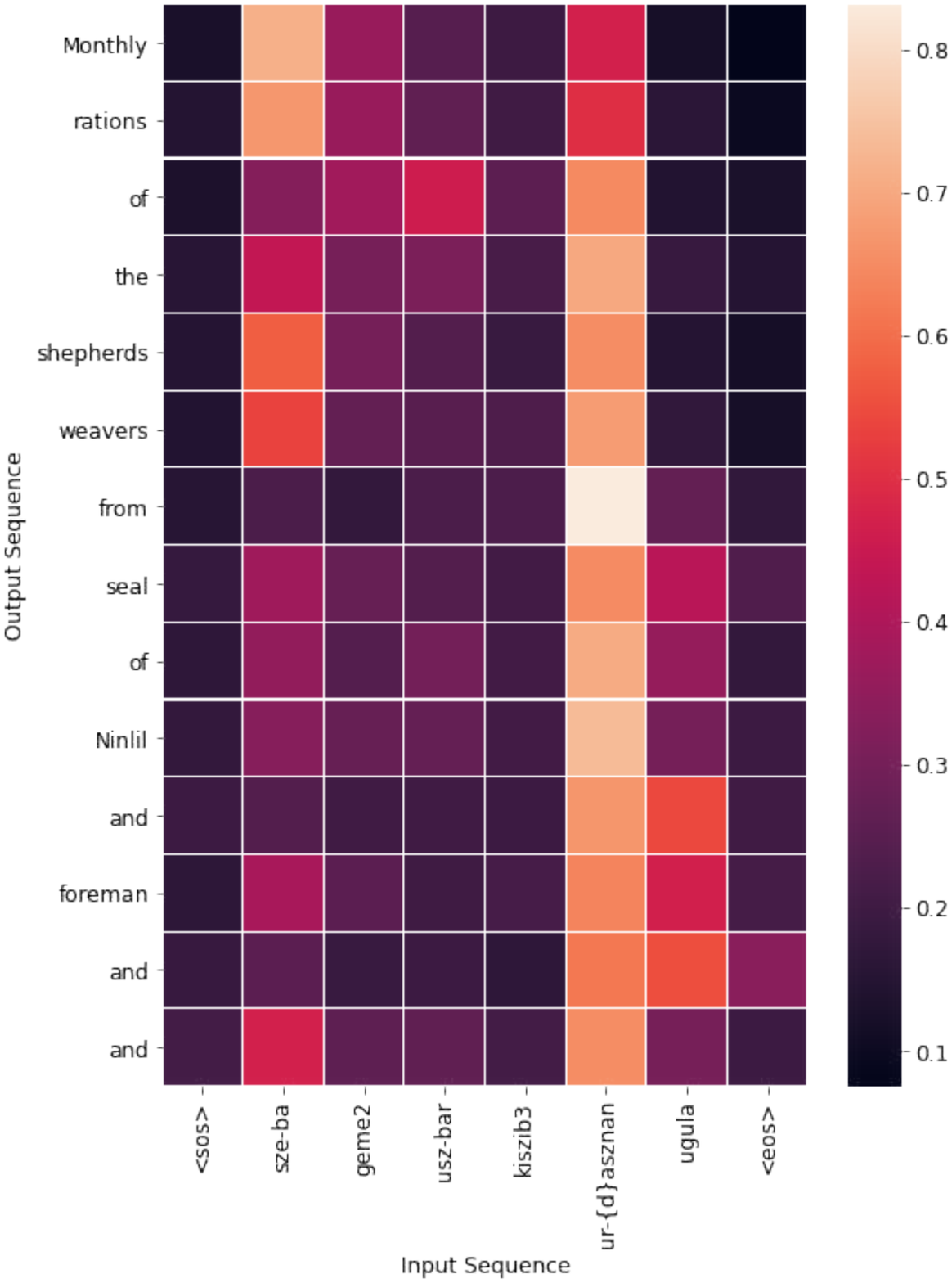}
          \label{fig:augsal}
        \end{subfigure}%
        \hspace{0.05cm}
        \begin{subfigure}{0.33\textwidth}
          \includegraphics[scale=0.25, trim=0cm 0cm 0cm 0cm]{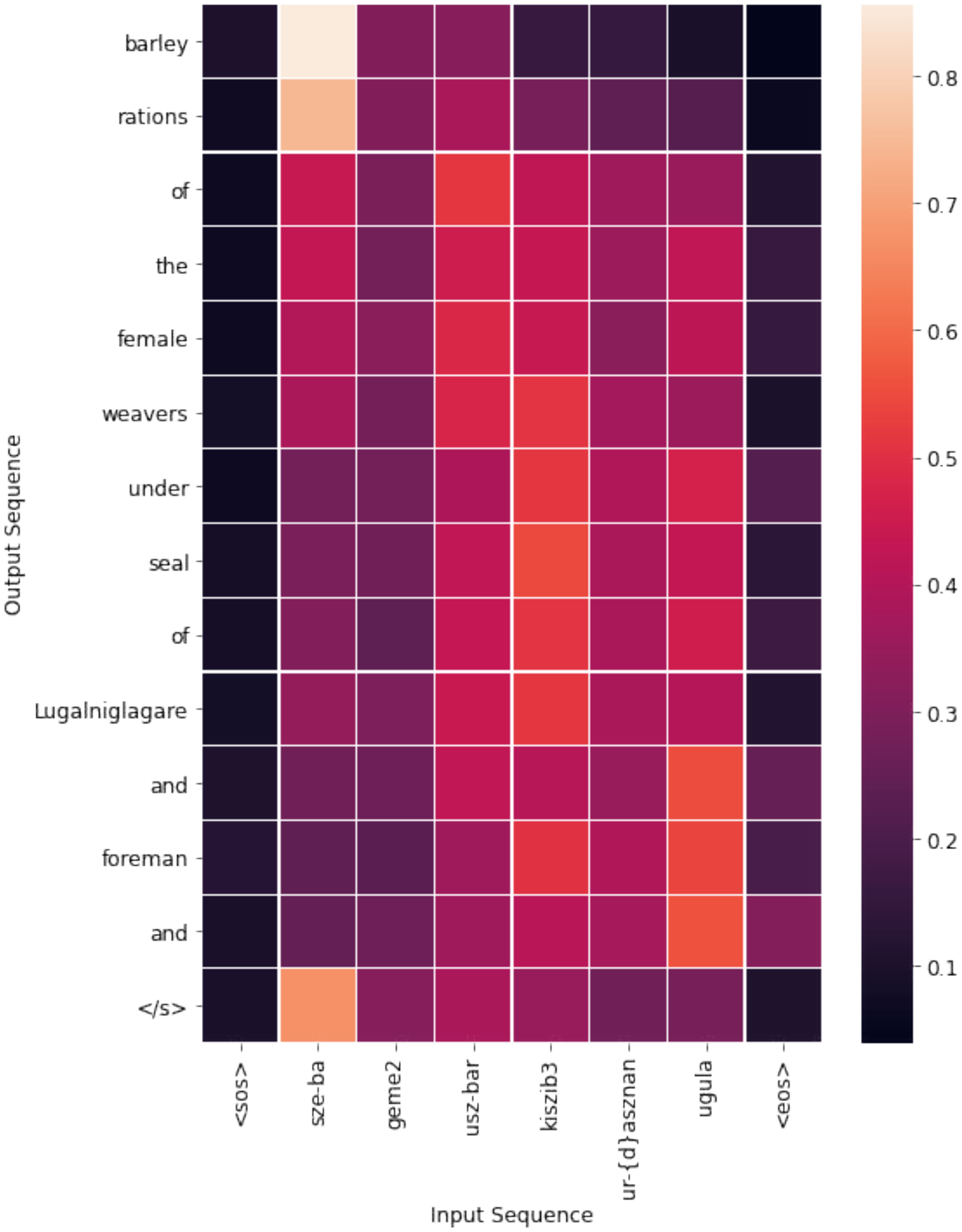}
          \label{fig:10xsal}
        \end{subfigure}%
    \vspace{-0.35cm}
    \caption{Grad-Norm Saliency}
    \label{Sal}
    \end{subfigure}
        \begin{subfigure}{\textwidth}
        \begin{subfigure}{0.33\textwidth}
          \includegraphics[scale=0.25, trim=0cm 0cm 0cm 0cm]{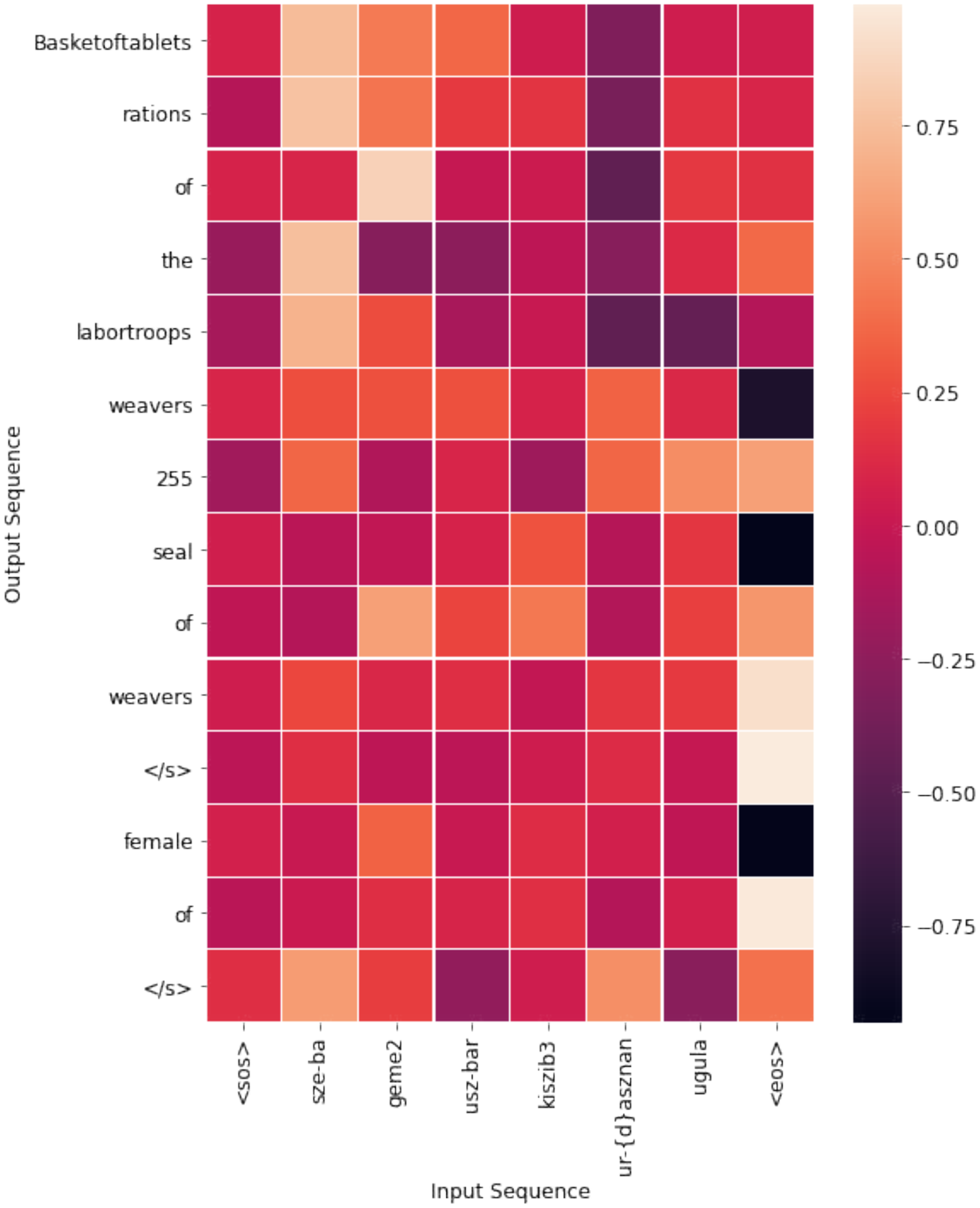}
          \label{fig:tlmig}
        \end{subfigure}%
        \hspace{0.05cm}
        \begin{subfigure}{0.33\textwidth}
          \includegraphics[scale=0.25, trim=0cm 0cm 0cm 0cm]{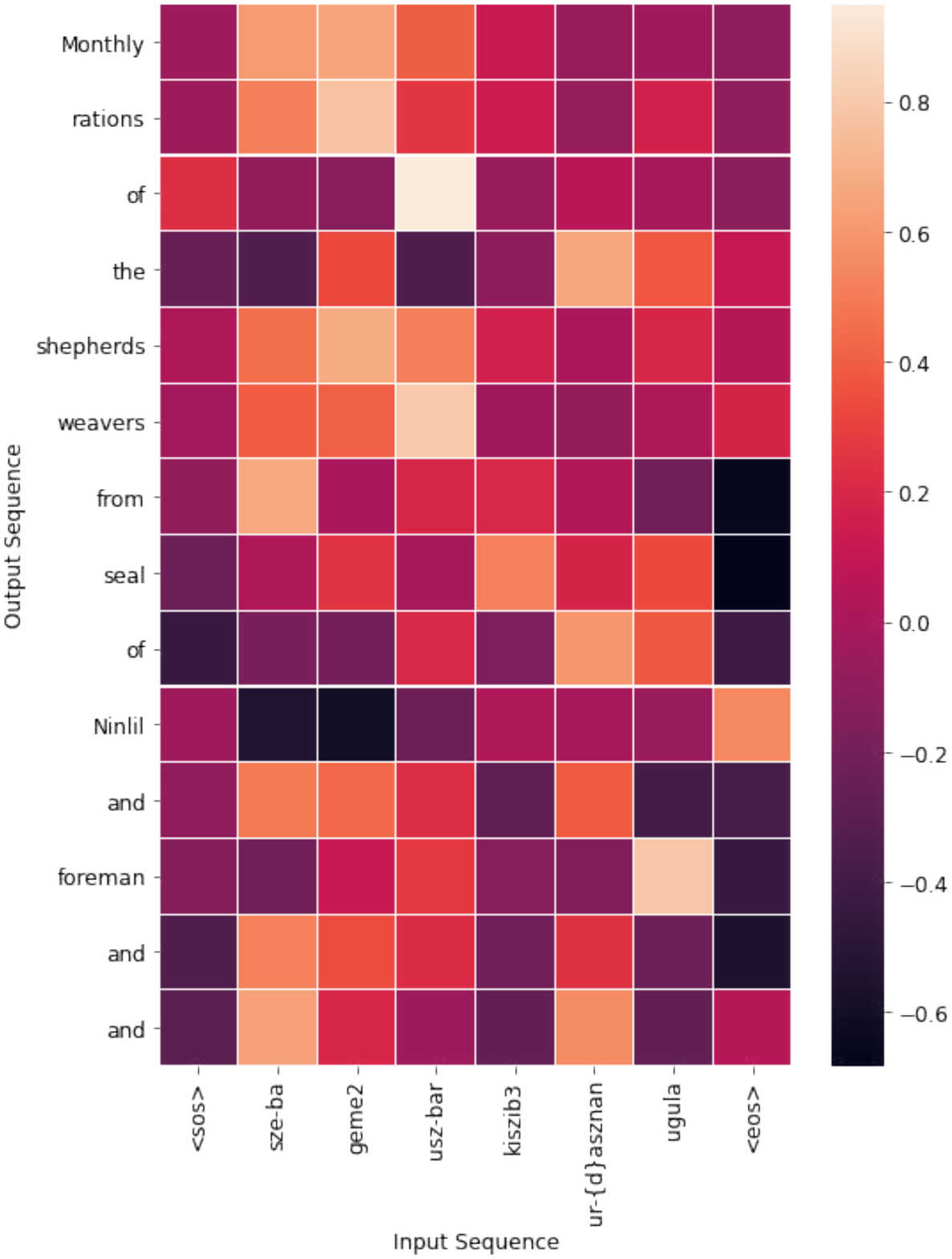}
          \label{fig:augig}
        \end{subfigure}%
        \hspace{0.05cm}
        \begin{subfigure}{0.33\textwidth}
          \includegraphics[scale=0.25, trim=0cm 0cm 0cm 0cm]{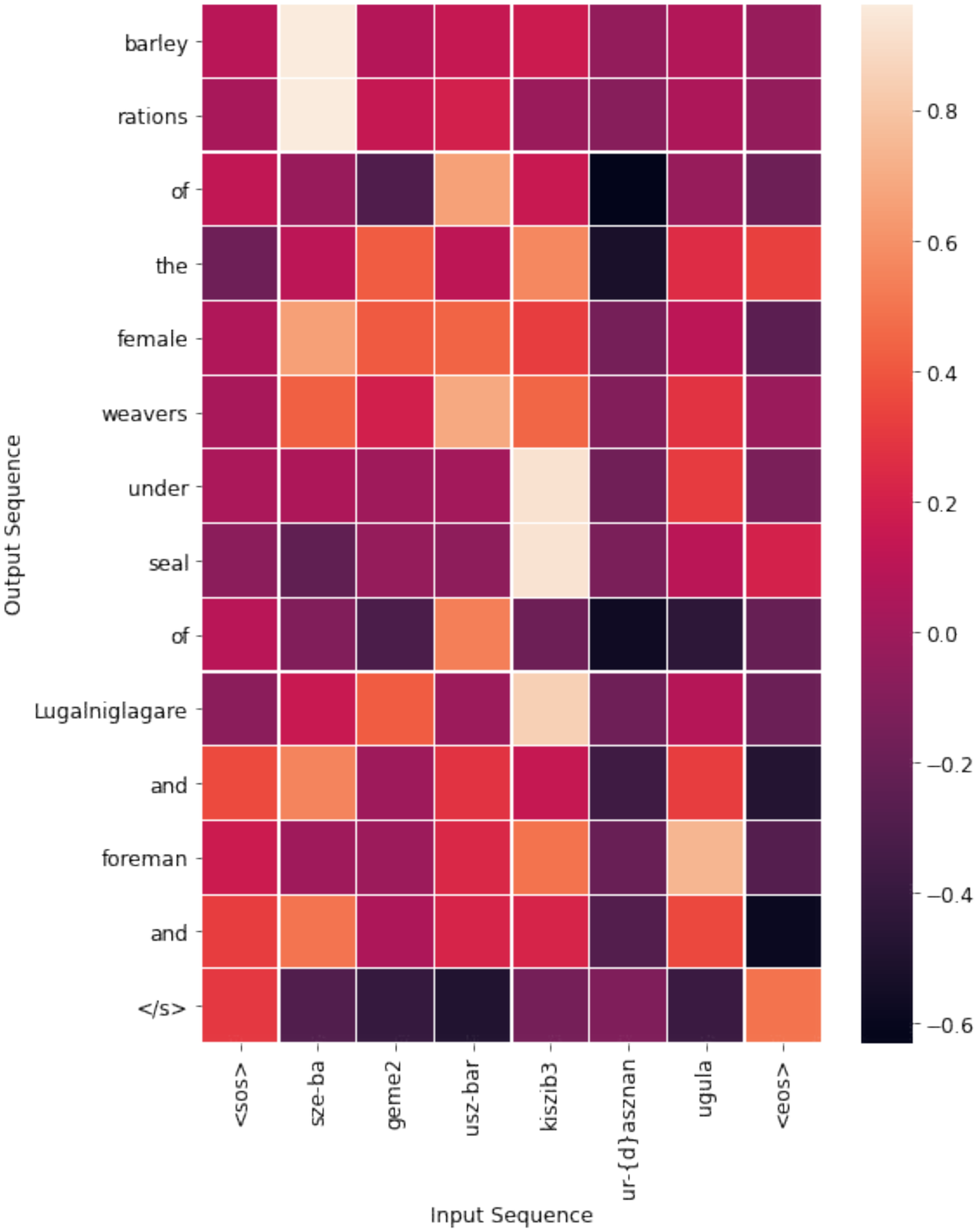}
          \label{fig:10xig}
        \end{subfigure}%
        \vspace{-0.35cm}
        \caption{Integrated Gradients}
    \label{IG}
    \end{subfigure}
    \label{GB}
    \caption{Comparing different gradient-based approaches used in \tname}
\end{figure*}

Figure \ref{Sal} represents the attribution heat-map when gradient-normalisation saliency (Simonyan, Vedaldi, and Zisserman 2013) is used. Being one of the most conventional techniques for finding attribution, it is more prone to inconsistent interpretations. Whereas, the attribution heat-map in Figure \ref{IG} represents the Integrated Gradients (IG) (Sundararajan, Taly, and Yan512017) approach. Being a path-based technique, which measures the gradient attribution relation using a straight-line path from a baseline (usually all-zeros), to the given input, it is much more robust and stable.

\begin{figure*}[htb]
    \vspace{-0.8cm}
    \begin{subfigure}{\textwidth}
        \begin{subfigure}{0.33\textwidth}
          \centering
          \includegraphics[scale=0.25, trim=0cm 0cm 0cm 0cm]{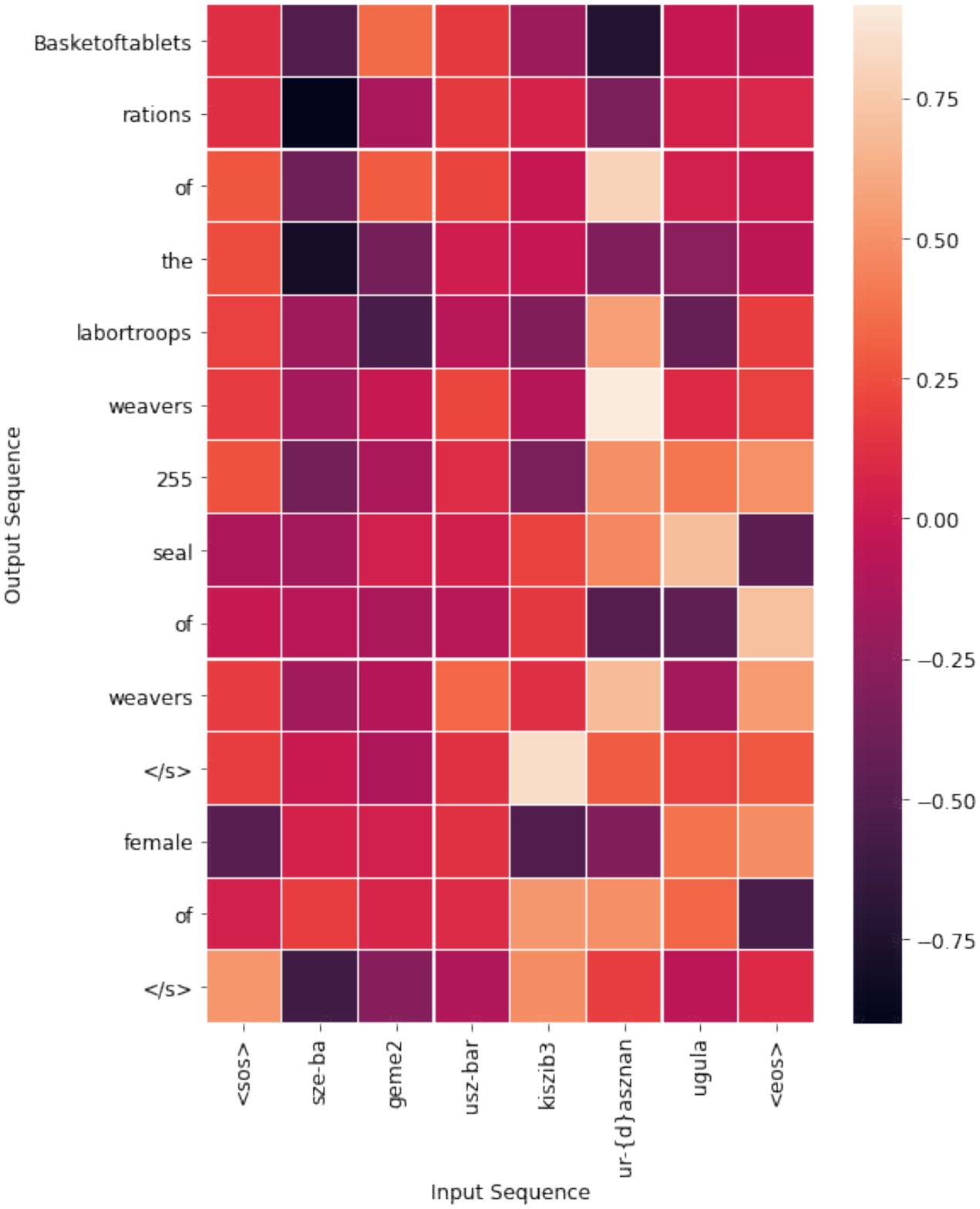}
          \label{fig:tlmfa}
        \end{subfigure}%
        \hspace{0.05cm}
        \begin{subfigure}{0.33\textwidth}
          \centering
          \includegraphics[scale=0.25, trim=0cm 0cm 0cm 0cm]{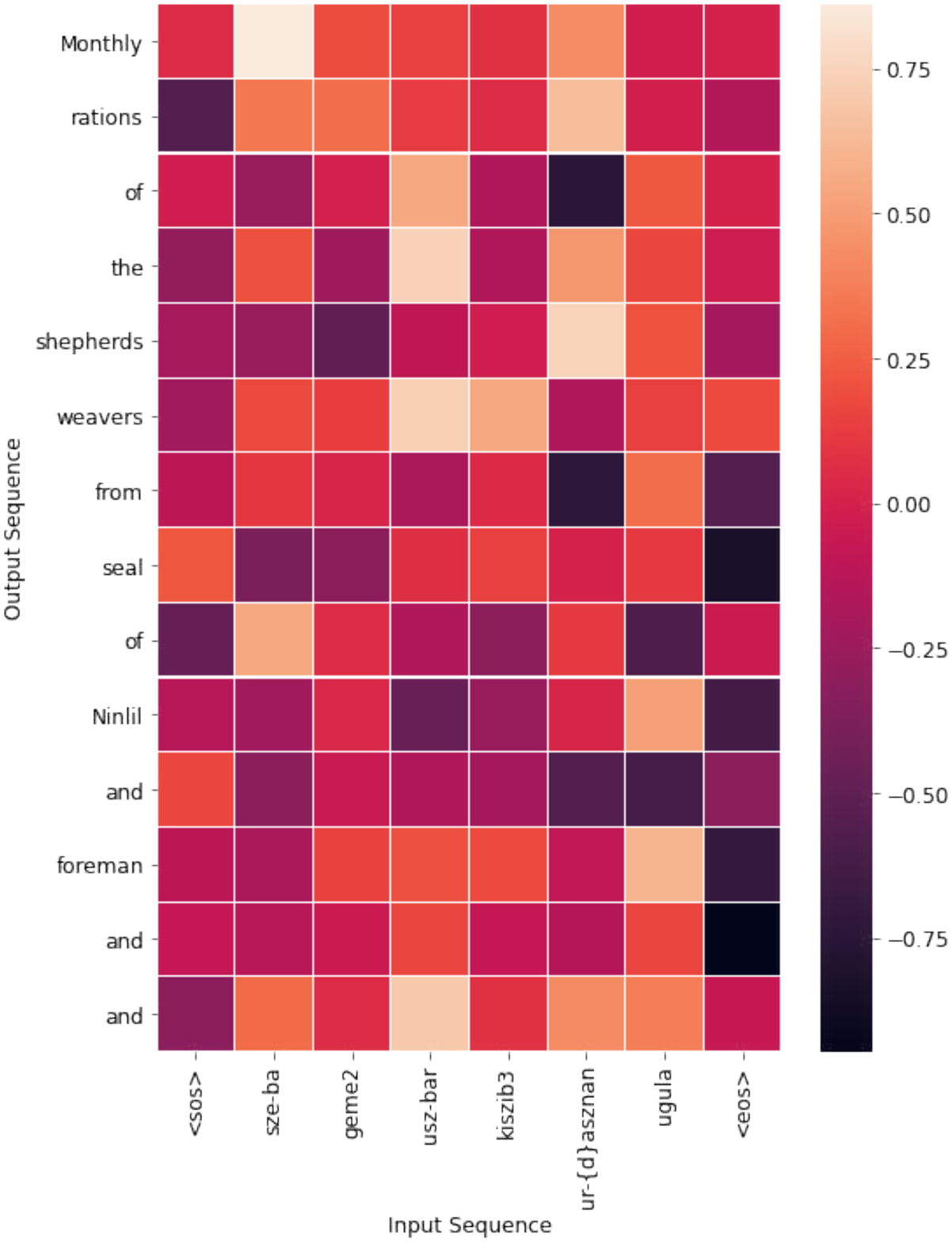}
          \label{fig:augfa}
        \end{subfigure}%
        \hspace{0.05cm}
        \begin{subfigure}{0.33\textwidth}
          \centering
          \includegraphics[scale=0.25, trim=0cm 0cm 0cm 0cm]{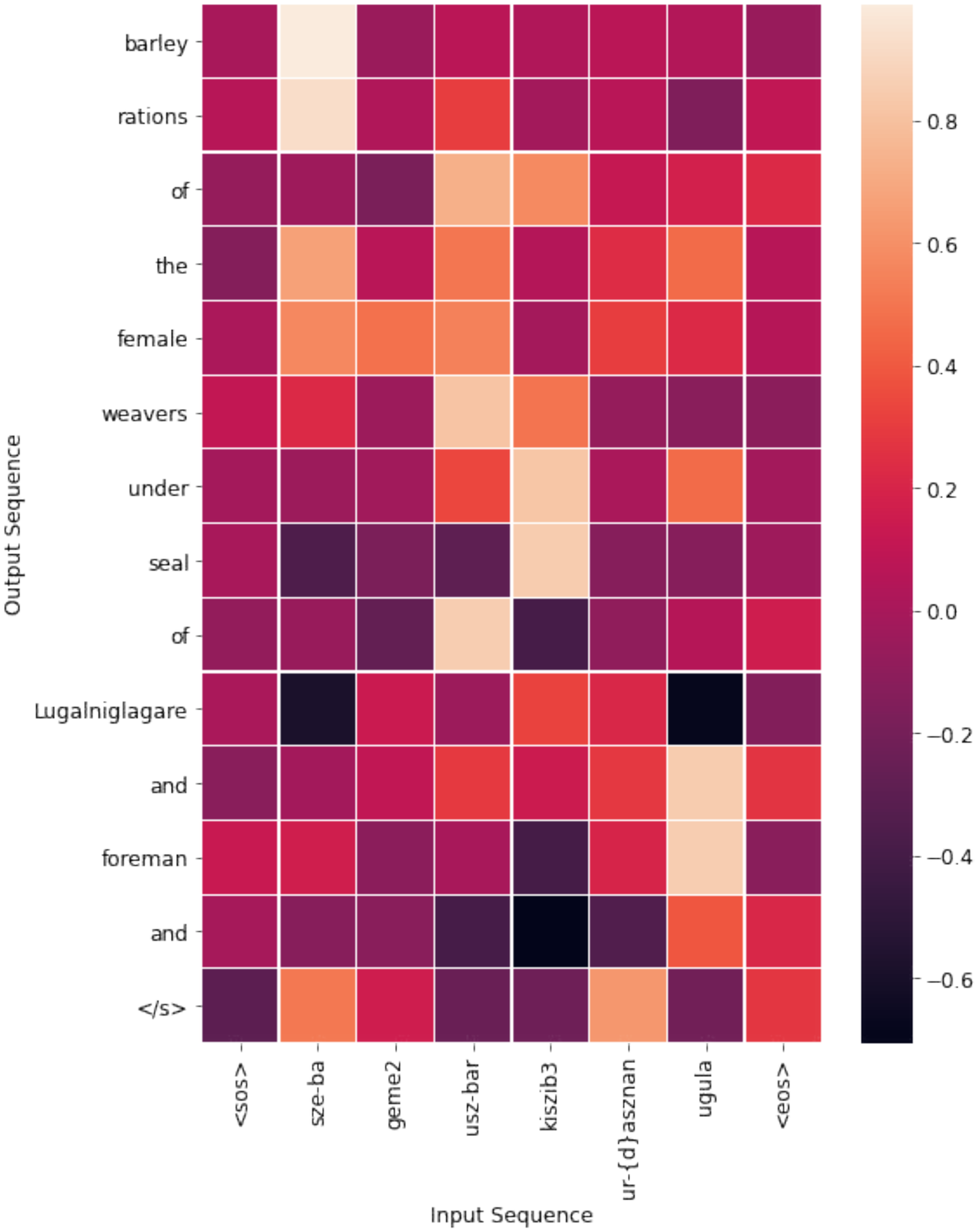}
          \label{fig:10xfa}
        \end{subfigure}%
    \end{subfigure}
    \label{FA}
    \vspace{-0.5cm}
    \caption{Feature Ablation in \tname}
\end{figure*}

Even though the gradient-based methods are much faster than perturbation-based methods, we observe that the heavy dependency of IG on hyper-parameters like the number of input steps to be considered when going from a baseline to the actual input, $n\_steps$, to be a major setback. The final attribution is generally found out after integrating (or summing) over the attributions of these sub-steps. We found that the attributions do not change when going beyond $n\_steps=250$, thus, we experiment by varying it between 10 to 250. We observe that there is no ideal value of $n\_steps$, IG's faithfulness to the model varies largely over this range. For some inputs, the best value is $n\_steps=50$ while for others $n\_steps=250$ is the most ideal. We judge this by considering how much the attribution is given to \textit{sos} and \textit{eos} tokens for each output token. Thus, based on both \textit{plausibility} and \textit{faithfulness}. We use $n\_steps=50$ for obtaining the heat-maps in Figure \ref{IG}.

\begin{figure*}[htb]
    \begin{subfigure}{0.49\textwidth}
        \includegraphics[width=\textwidth]{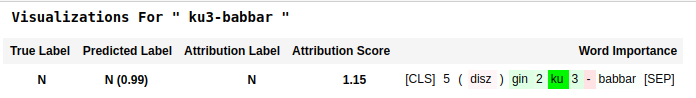}
        \includegraphics[width=\textwidth]{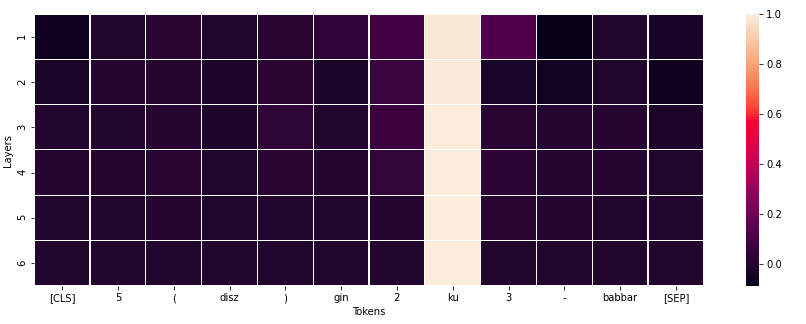}
        \includegraphics[width=\textwidth]{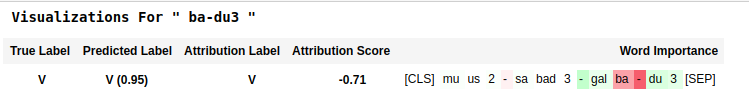}
        \includegraphics[width=\textwidth]{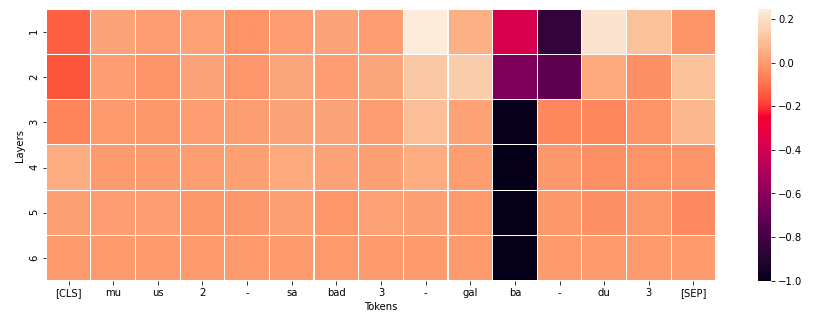}
        \caption{POS Tagging}
        \label{POSI}
    \end{subfigure}
    \begin{subfigure}{0.49\textwidth}
        \includegraphics[width=\textwidth]{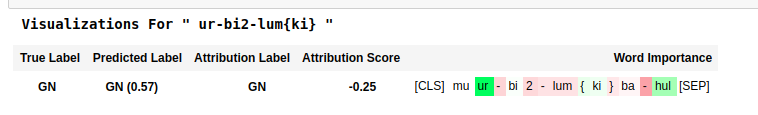}
        \includegraphics[width=\textwidth]{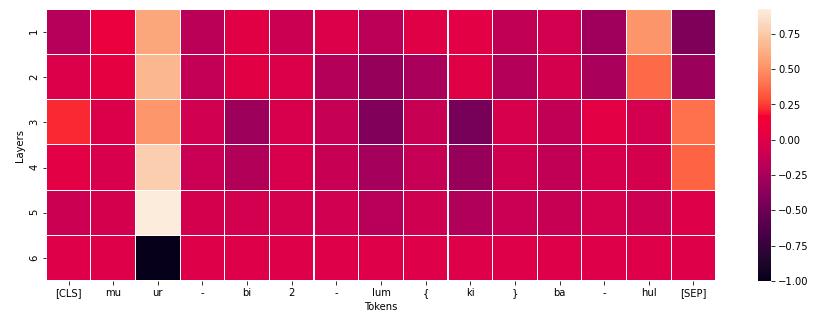}
        \includegraphics[width=\textwidth]{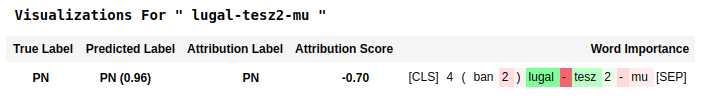}
        \includegraphics[width=\textwidth]{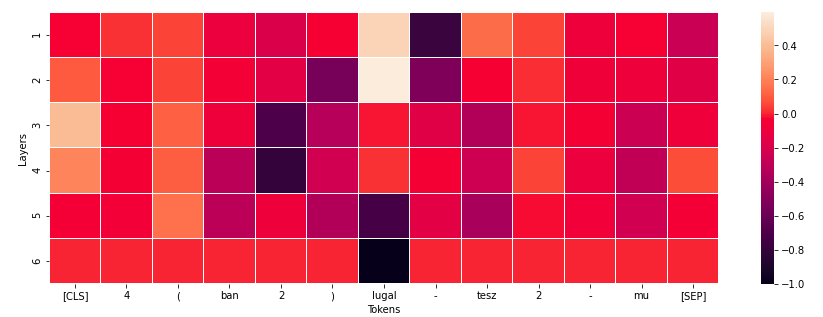}
        \caption{NER}
        \label{NERI}
    \end{subfigure}
    \caption{\tname\ on RoBERTa for Sequence Labeling}
    \label{sq}
\end{figure*}

Figure \ref{sq} represents the visualization for our sequence labeling tasks. It indicates two major things, 1) the effect of words, sub-words (depends on tokenization) on tagging the target word and 2) the effect of 6 transformer encoder layers. We created the hook on embeddings of RoBERTa with layer IG and obtained the visualizations for how each sub-word is contributing to tag the target word. Similarly, to obtain the heat-map we created the hook on RoBERTa embeddings and used the Layer Conductance. 

From Figure \ref{POSI} it can be observed that \textit{ku} and \textit{du} contribute the most to the attribution scores for tagging \textit{ku3-babbar} and \textit{ba-du3} as a Noun (N) and Verb (V), respectively. From the heat-maps it is also noted that \textit{ku} shows the effect on all 6 layers whereas in second example effects are majorly due to the initial transformer layers. Similarly in the Figure \ref{NERI} \textit{ur} and \textit{lugal} are the most effective sub-words to tag \textit{ur-bi2-lum{ki}} and \textit{lugal-tesz2-mu} as GN (Geographical Name) and PN (Personal Name) respectively. It is also interesting to note that both of these sub-words have a very positive impact in the initial layers but are contributing oppositely in the last layer.

\subsection{Human Evaluation}
The scoring by human experts was done independently for each result according to the following criterion:

• \textbf{3 (good)}: interpretable in the correct meaning by a native speaker of English; (almost) no incorrectly translated content word (e.g., tolerant against some errors in word order, but not in incorrect words).

• \textbf{2 (helpful)}: partially distorted, but interpretable with some context information (tolerant against errors in word order and against
incorrect function words).

• \textbf{1 (incorrect)}: contains incorrectly translated content words and/or is un-interpretable.

\section{Rules for POS Tagging and NER}
We used certain language-specific rules to assist CRF for the sequence labeling tasks. The rules were identified by human experts and some of them are as mentioned here:

    •  A word starting with ``ur-", ``lu2-", or ``dumu" is most likely to be a personal name.
    
    •  If a word is followed by ``mu'', then the next phrase denotes a year name.
    
    •  If a word is followed by ``iti", it denotes a month name.
    
    •  Words containing ``ki" are mostly associated with geographical names (GN).
    
    •  Words ending with part ``-hul" majorly denotes verbs.
    
    •  Words containing ``\{d\}" denotes either personal name (PN) or divine name (DN).
    
    •  A word followed by ``gin'' (\textit{unit}) majorly replicate a noun.


\end{document}